\documentclass{article}

\usepackage[main, final]{neurips_2026}
\usepackage{wrapfig}
\usepackage[utf8]{inputenc}
\usepackage[T1]{fontenc}
\usepackage[bookmarks=false]{hyperref}
\usepackage{url}
\usepackage{booktabs}
\usepackage{amsfonts}
\usepackage{nicefrac}
\usepackage{microtype}
\usepackage{xcolor}

\usepackage{graphicx}
\usepackage{subcaption}
\usepackage{amsmath,amssymb}
\usepackage{bm}
\usepackage{enumitem}
\usepackage{mathtools}
\usepackage{array}
\usepackage{multirow}
\usepackage{colortbl}
\usepackage{threeparttable}
\usepackage{fontawesome5}
\usepackage{listings}
\usepackage{amsfonts,amsthm}

\newcommand{\llmname}[1]{{\fontfamily{pcr}\selectfont {#1}}}

\allowdisplaybreaks
\usepackage{algorithm}
\usepackage{algorithmic}

\newcommand{\includefigmaybe}[2][]{%
  \IfFileExists{#2}{\includegraphics[#1]{#2}}{\fbox{\scriptsize Missing: \texttt{#2}}}%
}

\DeclareMathOperator{\prox}{prox}
\DeclareMathOperator{\diag}{diag}
\DeclareMathOperator*{\argmin}{arg\,min}

\newcommand{\Pbb}{\mathbb{P}}
\newcommand{\sigm}{\sigma}
\newcommand{\PhiG}{\Phi}

\newcommand{\E}{\mathbb{E}}

\theoremstyle{plain}
\newtheorem{theorem}{Theorem}[section]
\newtheorem{proposition}[theorem]{Proposition}
\newtheorem{lemma}[theorem]{Lemma}
\newtheorem{corollary}[theorem]{Corollary}

\theoremstyle{definition}

\theoremstyle{remark}

\begin{document}

\title{\textsc{Raptor}: Ridge-Adaptive Logistic Probes}
\author{
  Ziqi Gao,  Yaotian Zhu,  Qingcheng Zeng,  Xu Zhao,  Ziqing Wang,  Feng Ruan,  Kaize Ding \\
  Northwestern University \\
  \texttt{ziqigao2026@u.northwestern.edu} \quad
  \texttt{kaize.ding@northwestern.edu}
}

\maketitle

\begin{abstract}
Probing studies what information is encoded in a frozen LLM’s \textbf{layer representations} by training a lightweight predictor on top of it. Beyond analysis, probes are often used operationally in \emph{probe--then--steer} pipelines: a learned \textbf{concept vector} is extracted from a probe and then injected via additive activation steering by adding it to a layer representation during the forward pass. The effectiveness of this pipeline hinges on estimating concept vectors that are accurate, directionally stable under ablation, and inexpensive to obtain. 
Motivated by these desiderata, we propose \textsc{Raptor} (\textbf{R}idge-\textbf{A}da\textbf{pt}ive L\textbf{o}gistic P\textbf{r}obe), a simple $\ell_2$-regularized logistic probe whose validation-tuned ridge strength yields concept vectors from normalized weights. Across extensive experiments on instruction-tuned LLMs and human-written concept datasets, \textsc{Raptor} matches or exceeds strong baselines in accuracy while achieving competitive directional stability and substantially lower training cost; these quantitative results are supported by qualitative downstream steering demonstrations. 
Finally, using the Convex Gaussian Min--max Theorem (CGMT), we provide a mechanistic characterization of ridge logistic regression in an idealized Gaussian teacher--student model in the high-dimensional few-shot regime, explaining how penalty strength $\lambda$ mediates probe accuracy and concept-vector stability, yielding structural predictions that qualitatively align with trends observed on real LLM embeddings.
\end{abstract}
\section{Introduction}

\emph{Probing} trains lightweight predictors on frozen LLM layer representations to identify whether a target concept is decodable from those representations \citep{alain2016understanding,belinkov2017neural,hewitt2019structural,tenney2019bert}. 
Beyond diagnosis, probes are increasingly used operationally: the learned probe can guide downstream interventions.

One such downstream intervention is \emph{steering}, defined as the modulation of behavior at inference time without updating model weights \citep{dathathri2019pplm,krause-etal-2021-gedi-generative,liu-etal-2021-dexperts}.
Within this domain, we focus on \emph{additive activation steering}, a technique that modifies a layer's representation by injecting a learned direction \citep{turner2023activation,rimsky-etal-2024-steering}.
For an input sentence $x$ tokenized as $(t_1,\ldots,t_T)$, let $h_{\ell,T}\in\mathbb{R}^p$ denote the \textbf{layer representation} of the last token at layer $\ell\in\{1,\ldots,L\}$.
Consistent with standard probing practices, we treat $h_{\ell,T}$ as a sentence-level summary.
Additive activation steering intervenes by directly editing this representation: $h_{\ell,T} \leftarrow h_{\ell,T} + \alpha\, v_\ell ,$ where $v_\ell\in\mathbb{R}^p$ represents the \textbf{concept vector} for the target concept at layer $\ell$, and $\alpha\in\mathbb{R}$ controls the steering strength.
This approach is particularly advantageous as it is simple to implement and imposes negligible inference overhead.
However, its effectiveness is contingent upon the quality of $v_\ell$ (and $\alpha$); if the estimated concept vector is noisy or brittle, the resulting steering becomes unreliable.

\paragraph{What is a good probe?}
For probe--then--steer applications, a good probe must satisfy three requirements:
\textbf{(i) accuracy}, so that the concept is reliably decoded from layer representations;
\textbf{(ii) directional stability}, so that the learned vector remains reusable under small data perturbations rather than reflecting idiosyncrasies of a particular split \citep{hewitt-liang-2019-designing,pimentel2020information};
and \textbf{(iii) computational efficiency}, since probes are typically trained across many layers, concepts, and models.
These criteria define our evaluation lens.

Motivated by these desiderata, we propose \textbf{\textsc{Raptor}}: \textbf{Ridge-Adaptive Logistic Probes}.
For each tuple of model, layer, and concept, \textsc{Raptor} fits a validation-tuned $\ell_2$-regularized logistic probe on frozen layer representations and uses its normalized weight vector as the concept vector $v_\ell$.
The single ridge parameter $\lambda$ provides a simple mechanism for balancing predictive accuracy and directional stability, while the estimator remains computationally inexpensive. This minimalist design is especially useful for probe--then--steer pipelines, where higher probe accuracy alone does not necessarily yield a reliable intervention direction \citep{ravichander-etal-2021-probing,agarwal2025contextmatters,tan2024steeringreliability}, and overly elaborate estimators can make exhaustive layer and model sweeps expensive \citep{belinkov-2022-probing}.

LLM probing often lies in a high-dimensional few-shot regime, where the layer representation dimension $p$ can be comparable to, or even exceed, the number of labeled examples $n$.
In this regime, $\ell_2$ regularization is not merely a numerical stabilizer: the ridge strength $\lambda$ can directly shape both predictive accuracy and the stability of the learned direction.
We therefore analyze ridge logistic regression in a high-dimensional Gaussian teacher--student model with $n,p\to\infty$ and $n/p\to\delta$.
Using the Convex Gaussian Min--max Theorem (CGMT), we characterize how $\lambda$ controls the signal-aligned and orthogonal components of the learned direction, yielding structural predictions that qualitatively match trends on real LLM embeddings.

We evaluate \textsc{Raptor} through a benchmark spanning instruction-tuned models, human-written concept datasets, and full network depth.
Across accuracy, directional stability, and training cost, \textsc{Raptor} matches or exceeds representative alternatives while remaining simple to train.
We further include downstream steering evaluations showing that the learned directions support reliable additive control.

Our main contributions are as follows:
\begin{itemize}[leftmargin=*, itemsep=0pt, topsep=0pt]
    \item \textbf{Raptor algorithm.}
    We introduce a one-knob ridge-logistic probing pipeline with explicit hyperparameter selection of $\lambda$, producing interpretable concept vectors suitable for additive activation steering.
     \item \textbf{Operator-aligned evaluation criterion.}
    We formalize concept-vector estimation for additive activation steering and argue that probes should be evaluated under a joint objective capturing \emph{accuracy}, \emph{directional stability}, and \emph{computational cost}. 
    We instantiate this criterion through a multi-model, multi-dataset, multi-layer benchmark comparing \textsc{Raptor} against representative alternatives.
    \item \textbf{Theory-backed interpretation.}
    We provide a self-contained high-dimensional characterization of ridge logistic regression that explains how $\lambda$ mediates accuracy and concept-vector stability in the proportional regime.
\end{itemize}

\begin{figure*}[t]
   \includegraphics[width=\linewidth]{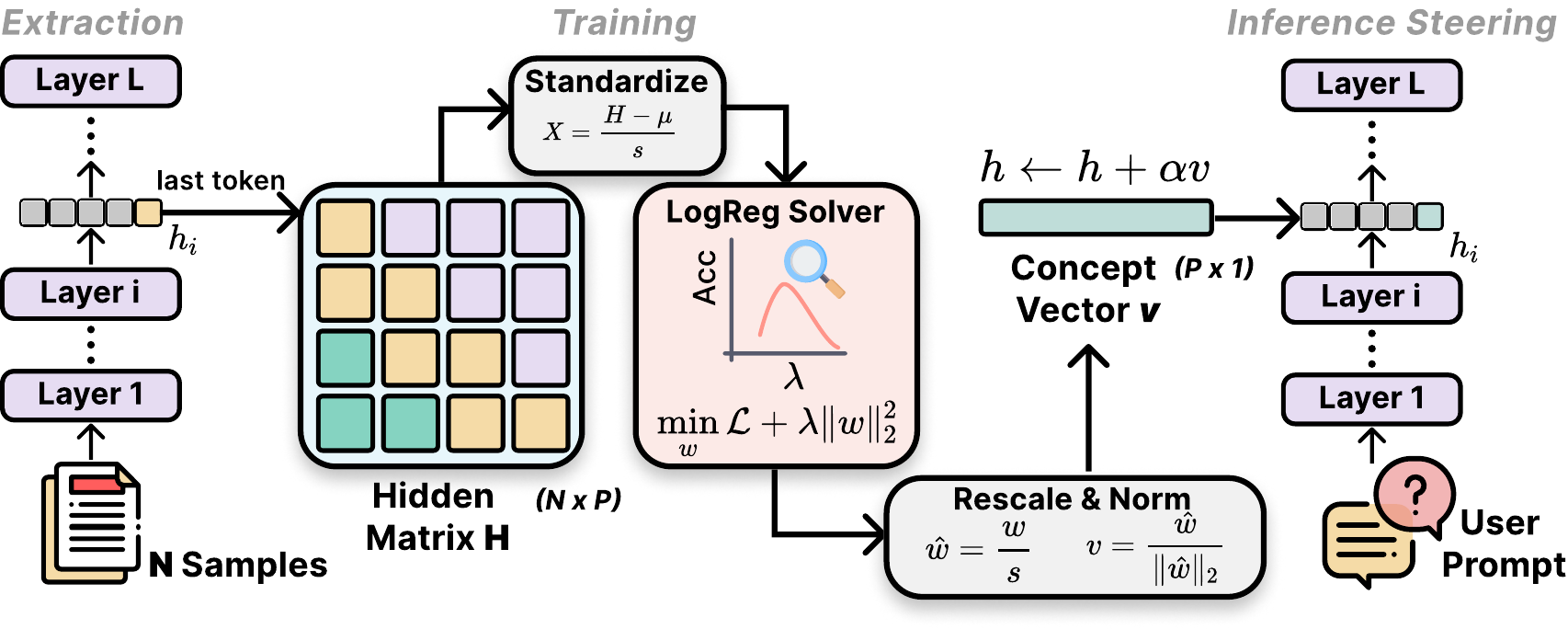}
   \vspace{-6mm}
\caption{\textbf{Overview of the \textsc{Raptor} pipeline for additive activation steering:} extract layerwise last-token embeddings, standardize features, fit an $\ell_2$-regularized logistic probe with $\lambda$ selected on a validation split, rescale to embedding space and normalize to obtain the concept vector $v_\ell$, then steer at inference by $h_{\ell,T}\leftarrow h_{\ell,T}+\alpha v_\ell$.}
   \label{fig:overview}
   \vspace{-3mm}
\end{figure*}

\section{Methodology}
\label{sec:method}

\subsection{Related work}
\label{sec:setup_related}

Probing studies what information is encoded in a model's \emph{layer representations} by training an auxiliary predictor on top of frozen activations \citep{alain2016understanding,belinkov2017neural,tenney2019bert,hewitt2019structural,belinkov-2022-probing}. 
In this work, probing is not only diagnostic but also operational: the learned probe is used to estimate a concept direction for additive activation steering.
For a concept $c$, model $M$, and layer $\ell$, we collect last-token layer representations
$\{h_i^{(\ell)}\}_{i=1}^N$, where $h_i^{(\ell)}\in\mathbb{R}^p$, together with binary labels $y_i\in\{0,1\}$.
The goal is to estimate a layer-wise concept vector $v^{(\ell)}\in\mathbb{R}^p$ that can be injected into the native hidden-state space through an intervention of the form
$h^{(\ell)} \leftarrow h^{(\ell)}+\alpha v^{(\ell)}$.

Existing probe estimators expose different tradeoffs for this purpose.
Linear probes are simple, interpretable, and directly yield a single direction, but can be brittle in high-dimensional, limited-data regimes.
Nonlinear probes can improve predictive accuracy, but they do not naturally produce a single reusable concept vector.
Related methods interpret probe weights as concept directions, as in TCAV \citep{kim2018tcav}, or use learned directions for editing and removal, including iterative nullspace-based methods \citep{ravfogel2020inlp} and amnesic removal \citep{elazar2021amnesic}.
Contrastive Activation Addition (CAA) uses the classical mean-difference approach to estimate steering directions \citep{rimsky-etal-2024-steering}.
More recent concept-estimation methods represent concepts as subspaces or distributions over directions, such as Gaussian Concept Subspace (GCS) \citep{zhao2024gcs} and estimators based on Recursive Feature Machines (RFMs) \citep{beaglehole2025rfm,zhao2024gcs}.

In probe--then--steer pipelines, however, the intervention form is already fixed: one adds a learned direction to an internal activation.
Thus the central methodological question is how to estimate a direction that is accurate, stable under small training perturbations, and inexpensive to obtain.
\textsc{Raptor} targets this regime with a one-knob ridge-logistic probe: it preserves the operational simplicity of a linear concept vector while using validation-tuned $\ell_2$ regularization to improve stability and optimization behavior.

\subsection{Ridge-Adaptive Logistic Probe}
\label{sec:Raptor}

Our choice of ridge logistic regression is motivated by the high-dimensional, small-sample regime typical of LLM probing: labeled hidden states are often nearly linearly separable, as shown empirically in Table~\ref{tab:svm_separability_nowarn}.
In this setting, the unregularized logistic loss may fail to attain
its infimum at finite parameter values. Under complete separation,
scaling a separating parameter pair $(w,b)$ sends its norm to
infinity while the loss decreases to zero.
This can lead to large-norm, poorly conditioned, and numerically unstable solutions.
When both classes are present in the training data, adding an
$\ell_2$ penalty to the weights restores a finite solution and
introduces a single interpretable regularization knob.
We therefore select the ridge strength $\lambda$ by validation for each model--layer--concept tuple.

\paragraph{Method setup.}
Fix a concept $c$, model $M$, and layer $\ell$.
We form a labeled dataset $\{(h_i^{(\ell)},y_i)\}_{i=1}^N$ with $h_i^{(\ell)}\in\mathbb{R}^p$ and $y_i\in\{0,1\}$, and use a fixed stratified split into training, validation, and test index sets
$\mathcal{I}_{\mathrm{tr}}$, $\mathcal{I}_{\mathrm{val}}$, and $\mathcal{I}_{\mathrm{te}}$.
Let $\tilde y_i=2y_i-1\in\{\pm1\}$.

For each layer, we standardize embeddings using training-set statistics only:
\begin{equation}
\mu_j^{(\ell)}
=
\frac{1}{|\mathcal{I}_{\mathrm{tr}}|}
\sum_{i\in\mathcal{I}_{\mathrm{tr}}} h_{ij}^{(\ell)},
\qquad
s_j^{(\ell)}
=
\sqrt{
\frac{1}{|\mathcal{I}_{\mathrm{tr}}|}
\sum_{i\in\mathcal{I}_{\mathrm{tr}}}
\big(h_{ij}^{(\ell)}-\mu_j^{(\ell)}\big)^2
},
\quad j=1,\dots,p .
\label{eq:standardization_stats}
\end{equation}
Zero entries of $s^{(\ell)}$ are replaced by $1$ in implementation.
We then form standardized features
\begin{equation}
x_{ij}^{(\ell)}
=
\frac{h_{ij}^{(\ell)}-\mu_j^{(\ell)}}{s_j^{(\ell)}} ,
\qquad j=1,\dots,p ,
\label{eq:standardized_features}
\end{equation}
and apply the same transform to validation and test embeddings.
This train-only standardization avoids information leakage and ensures that ridge tuning is not dominated by a small number of high-variance coordinates.

Given standardized features, \textsc{Raptor} fits ridge-regularized logistic regression with an intercept.
For an index set $\mathcal{I}\subseteq\{1,\dots,N\}$ and ridge strength $\lambda>0$, define
\begin{equation}
\mathcal{L}_{\lambda}^{(\ell)}(w,b;\mathcal{I})
=
\frac{1}{|\mathcal{I}|}
\sum_{i\in\mathcal{I}}
\log\!\Big(1+\exp\!\big(-\tilde y_i(w^\top x_i^{(\ell)}+b)\big)\Big)
+
\frac{\lambda}{2}\|w\|_2^2 ,
\label{eq:ridge_obj}
\end{equation}
where $(w,b)\in\mathbb{R}^p\times\mathbb{R}$.

We choose $\lambda$ on the validation split:
\begin{equation}
\lambda^\star
=
\arg\max_{\lambda\in\Lambda}
\mathrm{Acc}_{\mathrm{val}}(\lambda),
\label{eq:lambda_selection}
\end{equation}
then refit the probe on
$\mathcal{I}_{\mathrm{tr}}\cup\mathcal{I}_{\mathrm{val}}$
using the sample-size-adjusted ridge strength
$\lambda^\star
\frac{|\mathcal{I}_{\mathrm{tr}}|}
{|\mathcal{I}_{\mathrm{tr}}\cup\mathcal{I}_{\mathrm{val}}|}$.
Let the resulting standardized-space parameters be $(\hat w^{(\ell)},\hat b^{(\ell)})$.
To obtain a concept vector in the native embedding space, we fold the parameters back through the standardization map:
$
\hat\omega_j^{(\ell)}
=
\frac{\hat w_j^{(\ell)}}{s_j^{(\ell)}},
\
\hat b_{\mathrm{orig}}^{(\ell)}
=
\hat b^{(\ell)}-\langle \hat\omega^{(\ell)},\mu^{(\ell)}\rangle .
$
The original-space probe logit is therefore
$
    (\hat\omega^{(\ell)})^\top h^{(\ell)}+\hat b_{\mathrm{orig}}^{(\ell)}.
$
For additive activation steering, we use the normalized concept direction
$
v^{(\ell)}
=
\frac{\hat\omega^{(\ell)}}{\|\hat\omega^{(\ell)}\|_2}.$





\subsection{Steering Method}
\label{sec:gcav}

Given a learned layer-wise concept direction, additive steering only requires choosing the injection strength.
We adopt the \emph{GCAV} per-sample calibration rule~\cite{zhang2025gcav,xu2024uncovering} as an off-the-shelf way to set this strength without retraining. Fix a layer $\ell$ and let $h^{(\ell)}\in\mathbb{R}^p$ denote the (pre-steering) hidden activation.
We steer by $h^{(\ell)}_{\text{st}} \;=\; h^{(\ell)} + \alpha\, v^{(\ell)} .$ Along the direction $v^{(\ell)}$, the detector logit changes affinely:
\begin{equation}
(\omega^{(\ell)})^\top h^{(\ell)}_{\text{st}} + \beta^{(\ell)}
\;=\;
\underbrace{(\omega^{(\ell)})^\top h^{(\ell)} + \beta^{(\ell)}}_{\eqqcolon\, g^{(\ell)}(h)}
\;+\;
\alpha\,\|\omega^{(\ell)}\|_2 .
\end{equation}
Given a target probability level $p_0\in(0,1)$ and $g_0\coloneqq\mathrm{logit}(p_0)$, GCAV sets the \emph{minimal} strength that satisfies the target:
\begin{equation}
\begin{aligned}
\alpha_{\text{amplify}}^{(\ell)}(h)
\;=\;
\max\!\left\{0,\ \frac{g_0 - g^{(\ell)}(h)}{\|\omega^{(\ell)}\|_2}\right\}, \quad
\alpha_{\text{remove}}^{(\ell)}(h)
\;=\;
\min\!\left\{0,\ \frac{g_0 - g^{(\ell)}(h)}{\|\omega^{(\ell)}\|_2}\right\}
\end{aligned}
\end{equation}
In words, we inject only when the current activation does not already meet the target; the required strength is computed in closed form from the probe logit.

\definecolor{bestcolor}{HTML}{8443AD}
\definecolor{headergreen}{HTML}{85D3C1}
\definecolor{textgray}{gray}{0.45}
\definecolor{linegray}{gray}{0.65} 

\newcommand{\res}[2]{#1 \textcolor{textgray}{\scriptsize{(#2)}}}
\newcommand{\best}[2]{\cellcolor{bestcolor!15}\textbf{#1} \textcolor{textgray}{\scriptsize{(#2)}}}

\begin{table*}[t]
\centering
\caption{
\textbf{Probe accuracy (\%) across models and datasets.} 
Results are presented as \textbf{Average over layers} \textcolor{textgray}{\scriptsize{(Best Layer)}}.
All values are means over five seeds; best layers are selected using validation accuracy.
The \colorbox{bestcolor!20}{\textbf{purple background}} highlights the highest average-over-layers accuracy within each model and column, including all ties at the displayed precision. 
The vertical line separates individual task results from the overall average across all six tasks.
}
\label{tab:acc-summary}

\setlength{\aboverulesep}{0pt}
\setlength{\belowrulesep}{0pt}
\renewcommand{\arraystretch}{1.15} 
\setlength{\tabcolsep}{0pt} 

\resizebox{\textwidth}{!}{
\begin{threeparttable}
\begin{tabular}{
    @{\hspace{4pt}}l@{\hspace{8pt}}    
    c@{\hspace{12pt}}                  
    @{\hspace{4pt}}c@{\hspace{8pt}}    
    @{\hspace{4pt}}c@{\hspace{8pt}}    
    @{\hspace{4pt}}c@{\hspace{8pt}}    
    @{\hspace{4pt}}c@{\hspace{8pt}}    
    @{\hspace{4pt}}c@{\hspace{8pt}}    
    @{\hspace{4pt}}c@{\hspace{12pt}}   
    !{\color{linegray}\vrule width 1pt} 
    @{\hspace{12pt}}c                  
}
\toprule
\textbf{Model} & \textbf{Method} & \textbf{STSA} & \textbf{Cities} & \textbf{Common} & \textbf{Counterfact} & \textbf{HateXplain} & \textbf{Sarcasm} & \textbf{Average} \\
\midrule

\rowcolor{headergreen!25}
\multicolumn{9}{c}{\textit{Qwen 2.5 Series}} \\

\multirow{5}{*}{\llmname{Qwen-3B-Instruct}}
 & GCS                     & \res{83.5}{92.6} & \res{84.1}{99.4} & \res{62.5}{67.9} & \res{64.7}{77.6} & \res{66.0}{69.8} & \res{81.1}{87.9} & \res{73.6}{82.5} \\
 & xRFM                    & \res{87.8}{93.4} & \best{91.6}{99.5} & \best{69.0}{72.0} & \res{66.6}{75.8} & \res{70.3}{74.3} & \res{85.2}{91.0} & \best{78.4}{84.4} \\
 & Fixed-reg               & \res{84.6}{91.7} & \res{87.4}{99.6} & \res{61.7}{65.2} & \res{63.1}{73.4} & \res{68.7}{71.3} & \res{81.5}{88.9} & \res{74.5}{81.7} \\
 & CAA                     & \res{78.1}{91.4} & \res{84.3}{99.7} & \res{64.1}{68.6} & \res{62.9}{77.8} & \res{62.6}{68.5} & \res{74.2}{82.7} & \res{71.0}{81.4} \\
 & \textbf{Raptor (Ours)}  & \best{88.1}{93.7} & \res{88.1}{99.5} & \res{68.6}{71.6} & \best{68.8}{80.0} & \best{70.9}{74.1} & \best{85.3}{90.9} & \res{78.3}{85.0} \\
\midrule

\multirow{5}{*}{\llmname{Qwen-7B-Instruct}}
 & GCS                     & \res{87.3}{93.9} & \res{87.5}{99.5} & \res{66.0}{70.8} & \res{68.6}{80.0} & \res{69.2}{72.3} & \res{84.1}{89.4} & \res{77.1}{84.3} \\
 & xRFM                    & \res{89.8}{94.2} & \best{93.7}{99.3} & \best{69.6}{73.0} & \res{69.3}{80.3} & \res{72.4}{75.2} & \best{87.5}{92.3} & \best{80.4}{85.7} \\
 & Fixed-reg               & \res{87.8}{93.2} & \res{89.7}{99.3} & \res{63.3}{65.6} & \res{65.2}{75.2} & \res{67.7}{69.1} & \res{85.3}{90.9} & \res{76.5}{82.2} \\
 & CAA                     & \res{80.7}{92.0} & \res{86.6}{98.8} & \res{64.7}{69.7} & \res{65.6}{78.3} & \res{64.2}{70.0} & \res{78.3}{85.3} & \res{73.4}{82.3} \\
 & \textbf{Raptor (Ours)}  & \best{90.1}{94.6} & \res{90.9}{99.4} & \best{69.6}{73.4} & \best{71.3}{81.5} & \best{73.0}{75.2} & \best{87.5}{92.2} & \best{80.4}{86.0} \\
\midrule

\multirow{5}{*}{\llmname{Qwen-32B-Instruct}}
 & GCS                     & \res{88.0}{94.1} & \res{91.5}{99.5} & \res{68.0}{71.9} & \res{73.3}{82.8} & \res{70.8}{73.6} & \res{86.2}{91.5} & \res{79.6}{85.6} \\
 & xRFM                    & \res{90.4}{94.9} & \best{95.5}{98.9} & \best{71.4}{74.4} & \res{74.8}{82.2} & \res{73.4}{75.9} & \best{89.4}{93.8} & \best{82.5}{86.7} \\
 & Fixed-reg               & \res{89.1}{94.0} & \res{93.4}{99.2} & \res{65.4}{68.0} & \res{71.7}{81.8} & \res{66.3}{68.3} & \res{88.0}{92.8} & \res{79.0}{84.0} \\
 & CAA                     & \res{80.9}{89.4} & \res{90.1}{98.9} & \res{66.3}{70.3} & \res{67.7}{80.4} & \res{64.9}{70.2} & \res{81.1}{87.9} & \res{75.2}{82.8} \\
 & \textbf{Raptor (Ours)}  & \best{90.7}{94.8} & \res{93.5}{98.8} & \res{71.1}{74.5} & \best{76.2}{83.9} & \best{74.0}{75.9} & \res{89.3}{93.3} & \best{82.5}{86.9} \\

\rowcolor{headergreen!25}
\multicolumn{9}{c}{\textit{Gemma Series}} \\

\multirow{5}{*}{\llmname{Gemma-7B-it}}
 & GCS                     & \res{85.6}{91.9} & \res{87.4}{98.6} & \res{65.3}{70.2} & \res{68.1}{78.3} & \res{67.7}{71.0} & \res{81.3}{87.4} & \res{75.9}{82.9} \\
 & xRFM                    & \res{89.0}{93.3} & \best{93.1}{98.9} & \best{69.7}{72.8} & \res{69.3}{77.5} & \res{71.4}{75.2} & \best{85.1}{88.7} & \best{79.6}{84.4} \\
 & Fixed-reg               & \res{87.2}{91.9} & \res{88.9}{98.8} & \res{63.2}{65.8} & \res{65.3}{74.7} & \res{68.1}{70.2} & \res{82.6}{88.3} & \res{75.9}{81.6} \\
 & CAA                     & \res{83.9}{90.5} & \res{87.2}{98.4} & \res{65.8}{69.6} & \res{66.1}{77.6} & \res{62.5}{67.4} & \res{78.2}{84.3} & \res{73.9}{81.3} \\
 & \textbf{Raptor (Ours)}  & \best{89.2}{93.2} & \res{90.6}{98.8} & \res{69.5}{72.6} & \best{71.2}{80.7} & \best{71.8}{75.4} & \best{85.1}{89.6} & \best{79.6}{85.0} \\

\rowcolor{headergreen!25}
\multicolumn{9}{c}{\textit{Llama Series}} \\

\multirow{5}{*}{\llmname{Llama-3.1-8B-Instruct}}
 & GCS                     & \res{89.7}{93.7} & \res{92.7}{99.8} & \res{67.8}{71.5} & \res{75.1}{87.3} & \res{71.7}{73.6} & \res{87.0}{91.0} & \res{80.7}{86.1} \\
 & xRFM                    & \res{91.7}{94.6} & \best{96.8}{99.5} & \best{71.8}{74.3} & \res{78.9}{87.3} & \res{74.2}{76.7} & \best{90.1}{93.5} & \best{83.9}{87.7} \\
 & Fixed-reg               & \res{90.0}{92.9} & \res{93.7}{99.8} & \res{65.5}{67.4} & \res{76.1}{85.5} & \res{67.6}{69.8} & \res{88.8}{93.2} & \res{80.3}{84.8} \\
 & CAA                     & \res{83.8}{92.5} & \res{88.4}{97.9} & \res{66.9}{70.2} & \res{68.2}{81.2} & \res{66.6}{70.7} & \res{82.7}{87.2} & \res{76.1}{83.3} \\
 & \textbf{Raptor (Ours)}  & \best{91.9}{94.2} & \res{94.4}{99.5} & \res{71.4}{73.5} & \best{80.0}{88.6} & \best{75.0}{77.0} & \best{90.1}{93.4} & \res{83.8}{87.7} \\
\midrule
\multirow{5}{*}{\llmname{Llama-3.1-70B-Instruct}}
 & GCS                     & \res{91.1}{93.9} & \res{95.1}{99.7} & \res{70.9}{73.2} & \res{80.6}{88.9} & \res{75.0}{76.7} & \res{91.3}{94.9} & \res{84.0}{87.9} \\
 & xRFM                    & \res{93.0}{94.8} & \best{97.7}{99.5} & \best{73.7}{76.0} & \res{84.0}{90.6} & \res{76.8}{78.8} & \res{93.7}{96.5} & \best{86.5}{89.4} \\
 & Fixed-reg               & \res{92.0}{93.6} & \res{96.1}{99.9} & \res{68.7}{70.4} & \res{81.9}{89.6} & \res{71.0}{72.6} & \res{93.3}{96.0} & \res{83.9}{87.0} \\
 & CAA                     & \res{82.3}{90.1} & \res{93.1}{99.3} & \res{68.1}{70.7} & \res{71.2}{83.4} & \res{68.7}{72.9} & \res{88.0}{92.9} & \res{78.6}{84.9} \\
 & \textbf{Raptor (Ours)}  & \best{93.3}{94.7} & \res{96.5}{99.9} & \res{73.6}{74.6} & \best{84.6}{90.9} & \best{77.4}{79.0} & \best{93.8}{96.5} & \best{86.5}{89.3} \\
\bottomrule
\end{tabular}
\end{threeparttable}
}
\end{table*}

\section{Experiments}
\label{sec:exp}

We evaluate probe quality for probe--then--steer pipelines along three axes:
(i) \emph{classification accuracy} on real LLM embeddings,
(ii) \emph{directional stability} of the learned concept axis under small training perturbations, and
(iii) \emph{computational cost}.
All methods operate on the same layerwise embeddings and share the same data split within each seed.

\subsection{Setup}
\label{sec:exp_setup}

We benchmark \textsc{Raptor} against xRFM, GCS, logistic regression without regularization tuning (Fixed-reg), and a raw mean-difference baseline with a Euclidean midpoint readout (CAA) on six instruction-tuned LLMs from the Llama, Qwen, and Gemma families, and six human-written binary concept datasets:
STSA \citep{socher2013recursive}, Cities \citep{jin-etal-2025-exploring}, Common \citep{jin-etal-2025-exploring}, CounterFact \citep{meng2022locating}, HateXplain \citep{mathew2021hatexplain}, and Sarcasm \citep{misra2023sarcasm}.
For each model--dataset pair, we extract all-layer last-token embeddings and train probes independently per layer.
We use five stratified splits (seeds 42--46), each with approximately $20\%$ test, $16\%$ validation, and $64\%$ training data. Because concept predictability varies by layer, we report both average test accuracy over layers and test accuracy at the layer with the highest validation accuracy, averaged over the five seeds; sample standard deviations are reported in the appendix.
\textsc{Raptor} tunes $\lambda$ on validation data and reports refit test performance; Fixed-reg follows the same procedure without tuning, using $\lambda=1/|\mathcal{I}_{\mathrm{tr}}|$ before refitting. CAA re-estimates class means on $\mathcal{I}_{\mathrm{tr}}\cup\mathcal{I}_{\mathrm{val}}$; xRFM and GCS follow their original pipelines.

\subsection{Probe accuracy}
\label{sec:exp_acc}

\begin{figure}[!htb]
  \centering
  \includegraphics[width=\columnwidth]{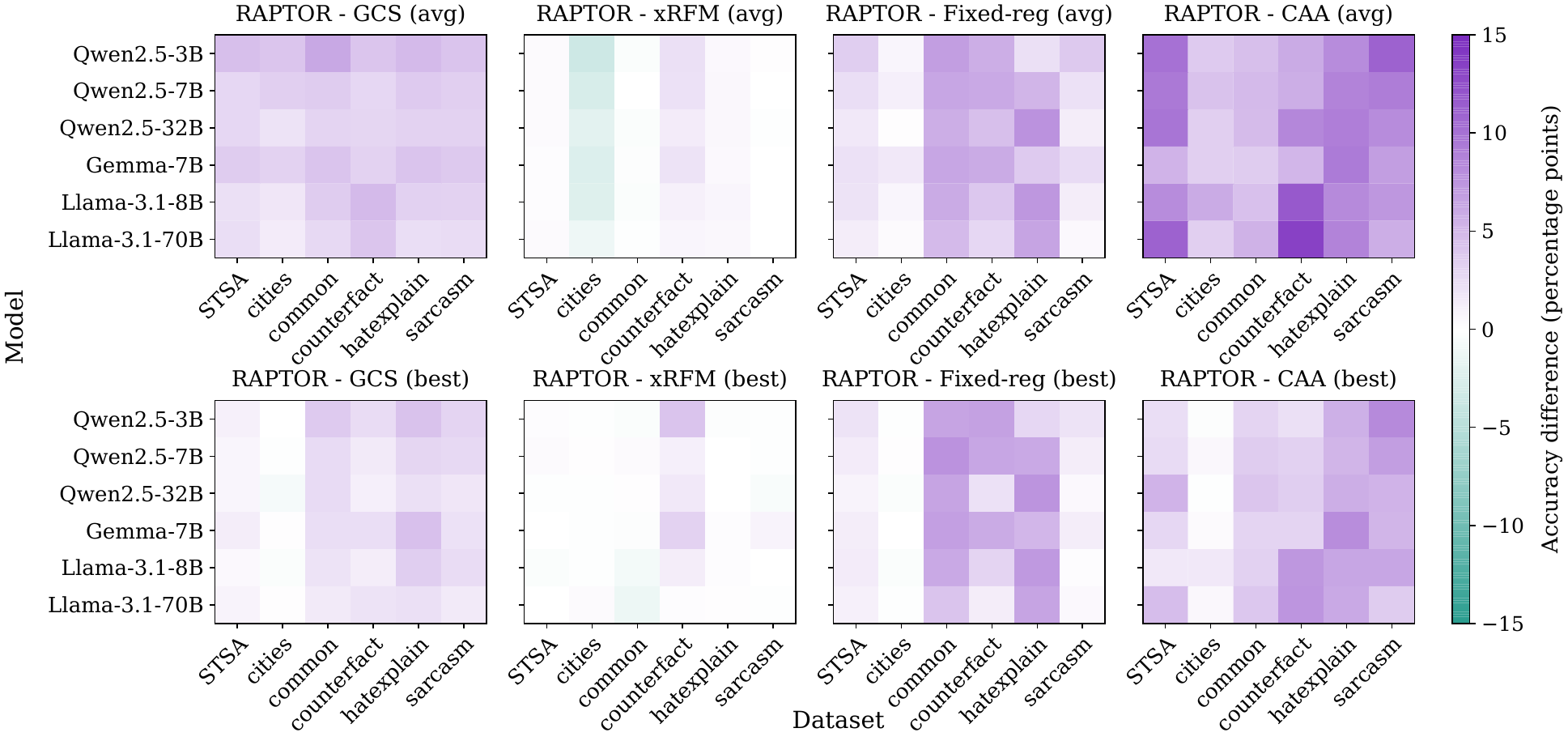}
   \vspace{-2mm}
  \caption{\textbf{Accuracy differences across the full model--dataset grid} (Raptor minus baseline) }
  \label{fig:acc_diff_2x2}
  \vspace{-7mm}
\end{figure}

Table~\ref{tab:acc-summary} reports probe accuracy over the full $6\times6$ model--dataset grid, with each cell showing layer-average test accuracy and validation-selected best-layer test accuracy in parentheses.
\textsc{Raptor} improves over GCS in all $36$ settings for layer-average accuracy and in $33/36$ settings for validation-selected best-layer accuracy.
Compared with xRFM, \textsc{Raptor} matches or exceeds validation-selected best-layer accuracy in $17/36$ settings ($16$ wins and $1$ tie), and improves layer-average accuracy in $21/36$ settings.
Averaged over the grid, \textsc{Raptor} reaches $86.64\%$ validation-selected best-layer accuracy, compared with $84.89\%$ for GCS (+$1.75$ percentage points) and $86.37\%$ for xRFM (+$0.28$ points). The corresponding layer-average accuracies for \textsc{Raptor} and xRFM are $81.85\%$ and $81.88\%$, respectively.
Compared with Fixed-reg and CAA, the mean gains are $3.52$ and $7.15$ percentage points for layer-average accuracy and $3.08$ and $3.96$ points for validation-selected best-layer accuracy, respectively.
On challenging semantic datasets such as HateXplain and Sarcasm, \textsc{Raptor} consistently achieves higher mean best-layer accuracy than GCS, Fixed-reg, and CAA across all six models.
Figure~\ref{fig:acc_diff_2x2} visualizes per-setting differences (\textsc{Raptor}$-$baseline), where purple indicates improvements and green indicates decreases.
For completeness, Algorithm~\ref{alg:Raptor_impl} provides detailed pseudocode and the full training/validation protocol; code and experiment scripts will be released upon publication.

\subsection{Directional robustness}
\label{sec:exp_robust}

We evaluate whether each method produces a stable concept axis under small perturbations of the concept-labeled training data.
We evaluate 18 model--dataset settings spanning Qwen2.5-3B-Instruct, Qwen2.5-32B-Instruct, and Gemma-7B-it across all six datasets, covering 768 layers.
For each (model, dataset, layer, method), we perform $K=20$ ablation runs.
In each run, we randomly drop $20\%$ of the samples in $\mathcal{I}_{\mathrm{tr}}\cup\mathcal{I}_{\mathrm{val}}$, re-split the remainder into training and validation sets with validation fraction $0.2$, and re-estimate the concept directions.
\textsc{Raptor} re-selects $\lambda$ on each ablated validation split, whereas Fixed-reg uses a predeclared regularization setting in every ablation. xRFM and GCS follow their original pipelines, while CAA re-estimates the raw mean-difference direction on the retained pool.

Let $\{v_r\}_{r=1}^K$ be the unit-normalized directions, excluding intercepts, returned across ablations for a fixed (model, dataset, layer).
We measure axis stability irrespective of sign using mean absolute pairwise cosine similarity: $\mathrm{Robust}
\;=\;
\frac{2}{K(K-1)}
\sum_{1\le r < s \le K}
\left|\left\langle v_r, v_s\right\rangle\right|.$
 We report the mean and maximum of this score across layers in \autoref{fig:robust_mean_max}, with diamonds indicating medians across the 18 settings. xRFM retains standardized-space directions, whereas the other methods use raw embedding coordinates. The top and bottom panels show maximum and mean robustness across layers, respectively. Each circle represents one model--dataset setting.
 
\par\noindent
\begin{minipage}[t]{0.58\textwidth}
    \vspace{0pt}
    \centering
    \captionsetup{type=figure}
    \includegraphics[width=\linewidth]
    {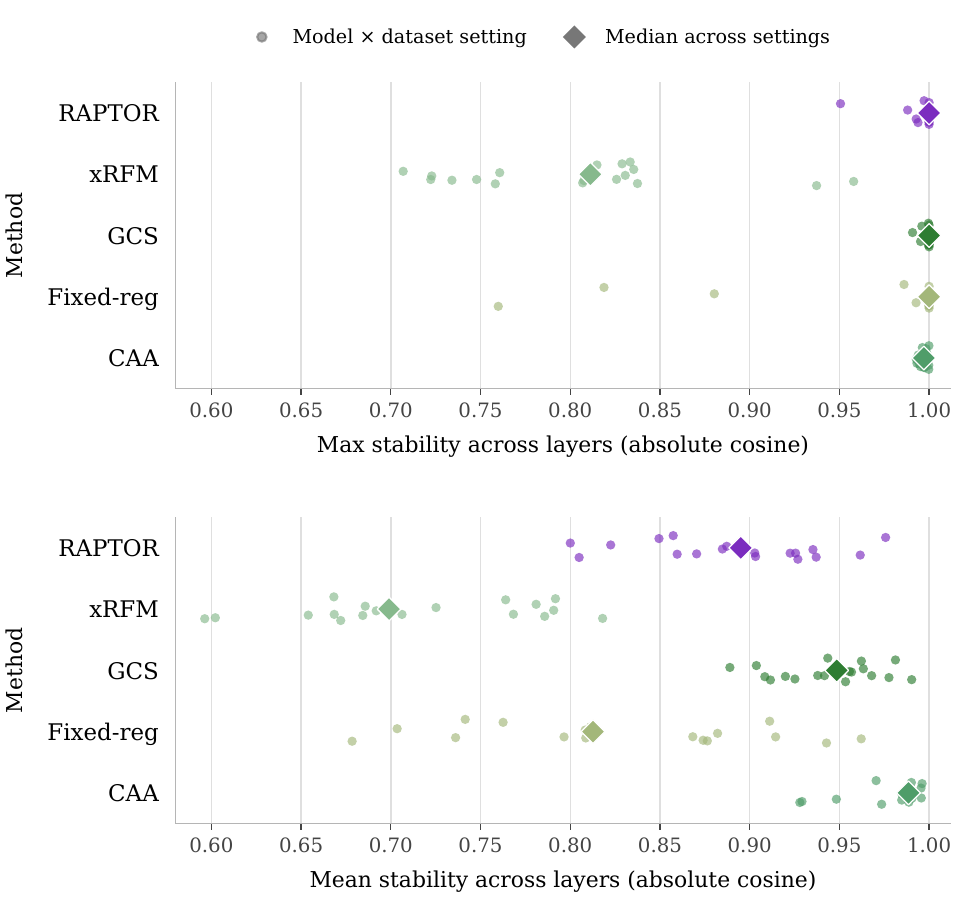}
    \caption{Directional stability.}
    \label{fig:robust_mean_max}
\end{minipage}\hfill
\begin{minipage}[t]{0.39\textwidth}
    \vspace{0pt}

    \par\smallskip
    Across the evaluated settings, \textsc{Raptor} consistently
    improves over xRFM. GCS and CAA remain more stable on average across layers,
    consistent with their use of averaging, but \textsc{Raptor}
    is close in maximum-layer robustness (the median maxima
    across the 18 settings round to $1.00$ for all three methods).
    Compared with Fixed-reg, \textsc{Raptor} also has higher
    median layer-average robustness ($0.90$ versus $0.81$). At the individual-setting level, \textsc{Raptor} achieves higher
layer-average robustness than Fixed-reg in $15$ of the $18$ settings.
This advantage holds for a majority of datasets within each of
the three evaluated models.
    Thus, \textsc{Raptor} yields concept axes that are less
    sensitive to finite-sample perturbations than xRFM while
    retaining high predictive accuracy, a prerequisite for
    directional stability to be practically useful.

\end{minipage}

\subsection{Computational cost}
\label{sec:exp_cost}

\begin{figure}[!htbp]
    \centering
    \includegraphics[width=0.9\linewidth]
    {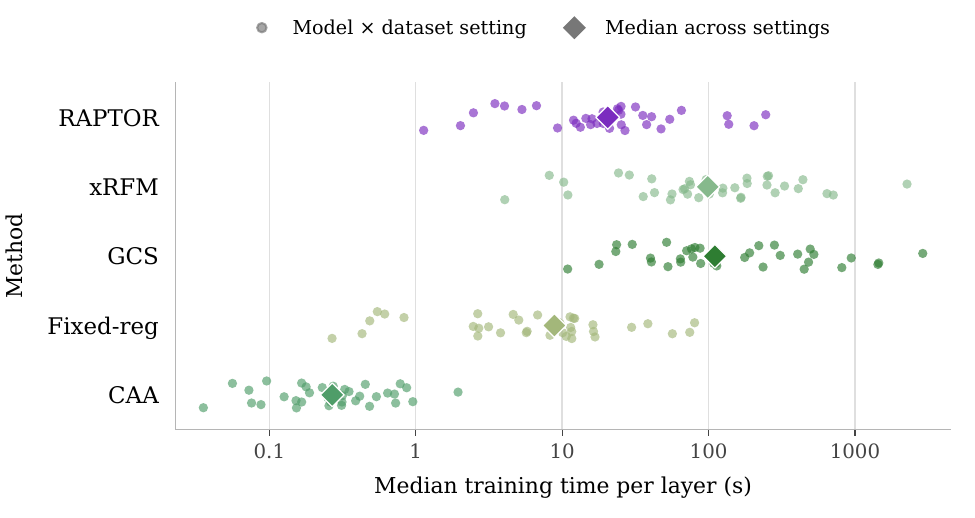}
    \caption{\textbf{Per-layer probe training time across
$6\times6$ model--dataset settings.}
The horizontal axis uses a logarithmic scale.
Each circle shows the per-layer training-time summary
for one model--dataset setting.
Diamonds indicate medians across the $36$ settings
for each method.}
    \label{fig:cost_dumbbell}
\end{figure}

We measure training cost as wall-clock time under the same hardware and data pipeline.
Each model--dataset--layer probe fit used one CPU worker.
Since probes are trained independently per layer, total fitting cost increases with the number of evaluated layers.
Across the full grid, \textsc{Raptor} is consistently faster than xRFM and GCS.
Figure~\ref{fig:cost_dumbbell} summarizes per-layer training time for all five methods across the same $36$ model--dataset settings.

For each (dataset, model) pair, we plot the per-layer time summary for each method, with diamonds indicating medians across settings.
CAA is particularly fast because its direction and midpoint readout are computed directly from class-conditional embedding means, without fitting a regression model.
Fixed-reg is also fast because it fits logistic regression with a preset regularization setting, avoiding the penalty-parameter search used by \textsc{Raptor}. Although \textsc{Raptor} is slower than CAA and Fixed-reg, their lower accuracy on semantically demanding datasets (\autoref{sec:exp_acc}) highlights the importance of predictive accuracy as the primary criterion for evaluating probes.
\textsc{Raptor} nevertheless remains substantially faster than GCS and xRFM across the evaluated settings.

\subsection{Steering results}
\begin{wraptable}{r}{0.50\textwidth}
\vspace{-1.0em}
\centering
\caption{\textbf{Steering controllability.}}
\label{tab:steering_control}
\small
\setlength{\tabcolsep}{3pt}
\begin{tabular}{@{}llcccc@{}}
\toprule
Dataset & Dir. & Succ. & Intv. & $|\alpha|$ med & $|\alpha|$ p90 \\
\midrule
CounterFact & away    & 1.000 & 0.833 & 10.50 & 66.50 \\
CounterFact & towards & 1.000 & 0.556 &  6.88 & 59.25 \\
HateXplain  & towards & 1.000 & 0.654 &  6.80 & 49.00 \\
HateXplain  & away    & 1.000 & 0.538 &  4.70 & 24.50 \\
Sarcasm     & towards & 1.000 & 0.715 &  8.83 & 28.00 \\
Sarcasm     & away    & 1.000 & 0.631 &  6.20 & 30.75 \\
STSA        & away    & 1.000 & 0.615 &  3.57 & 21.25 \\
STSA        & towards & 1.000 & 0.792 & 12.38 & 29.25 \\
\bottomrule
\end{tabular}
\label{tab:steer-control-main}
\vspace{-1.2em}
\end{wraptable}
We evaluate steering control using the concept vectors learned by our probes.
For a given concept and a target direction (either towards the concept or away from it), we intervene on the model by adding a scaled concept vector to the layers' representation.
We use an adaptive per-layer steering strength $\alpha_\ell$ chosen to drive the probe probability $p_m$ toward an extreme target (towards: $p_m \approx 0.9999$; away: $P_m \approx 0.0001$).
To avoid intervening with poorly aligned directions when the probe is unreliable, we optionally skip layers whose probe validation accuracy  falls below a reliability threshold $\tau$(e.g., $\tau=0.7$ for Counterfact and $\tau=0.8$ for STSA-positive in our runs).

We report three control metrics:
(i) \textbf{probe-target success rate}, the fraction of evaluated layer--prompt pairs whose final probe probability lands in the desired extreme region (towards: $P_m \ge 0.9999$; away: $P_m \le 0.0001$);
(ii) \textbf{intervention rate}, the fraction of evaluated layer--prompt pairs where steering is actually applied (i.e., the baseline is not already in the target region), and
(iii) \textbf{steering strength}, summarized by the distribution of $|\alpha_\ell|$ (median/p90/max).

\autoref{tab:steer-control-main} shows adaptive steering achieves near-perfect probe-coordinate control across all datasets and directions, while the required intervention rate varies by task (roughly $0.54$ to $0.83$ of pairs).
The steering strength exhibits a pronounced long tail: although median $|\alpha|$ is modest (about $3.6$ to $12.4$), some settings require very large interventions (max $|\alpha|$ up to $249$), with a markedly heavier tail for away-direction control.
We further break down the 'hard' layers that dominate the tail in Appendix~\ref{app:steer-layers}, and provide detailed pseudocode for our steering-strength selection procedure in Algorithm~\ref{alg:steer_towards}.

Because probe-coordinate success does not by itself guarantee downstream generation quality, we also conduct a blinded LLM-as-judge evaluation of open-ended steering outputs in Appendix~\ref{app:llm_steering_eval}. We use 100 frozen neutral movie prompts with three random seeds, yielding 300 prompt--seed cases. The judge first filters invalid outputs and then scores valid generations on target sentiment strength, output quality, naturalness, and overall steering success. We compare \textsc{Raptor} with GCS under the same GCAV-based strength calibration; xRFM is excluded because it is nonlinear and does not support the same closed-form calibration. On the fully valid subset, \textsc{Raptor} achieves stronger target sentiment control than GCS while maintaining comparable output quality and naturalness.
\vspace{-4mm}

\section{Mechanistic Analysis}
\label{sec:theory}
\paragraph{Motivation.}
Few-shot probing often operates in a proportional high-dimensional regime where the layer representation dimension $p$ can be comparable to, or exceed, the number of labeled examples $n$.
In this regime, empirical risk minimization may enter a nearly separable phase, unregularized logistic regression can become ill-posed, and classical fixed-$p$ theory can be inaccurate \citep{sur2019modern,montanari2025maxmargin,deng2021doubledescent}.
Ridge-regularized logistic regression is therefore a natural default: it restores well-posedness and introduces a single interpretable knob $\lambda$ whose effect can be characterized sharply using tools from precise high-dimensional analysis, including AMP and CGMT-based frameworks \citep{donoho2016highdimensional,thrampoulidis2015regularized,thrampoulidis2018mestimators,salehi2019impact}.
Mechanistically, $\lambda$ controls the decomposition of the learned direction into signal-aligned and orthogonal components, motivating our use of $\lambda$ as a unified control parameter for accuracy and directional stability.
\paragraph{Main results.}
We analyze ridge logistic regression under a Gaussian teacher--student model in the proportional regime $n,p\to\infty$ with $n/p\to\delta$.
Using CGMT \citep{thrampoulidis2015regularized,thrampoulidis2018mestimators}, we obtain a deterministic fixed-point characterization in terms of order parameters $(\bar\alpha,\bar\sigma)$ and an auxiliary scalar $\bar\gamma$.
This yields a scalar expression for the limiting test accuracy and clarifies how $\lambda$ shapes the signal--orthogonal decomposition of the learned direction.
\autoref{app:cgmt} gives a self-contained derivation; after matching notation, our fixed-point equations coincide with the ridge-logistic specialization of \cite{salehi2019impact}.

The Gaussian teacher--student model is not intended as an exact generative model for LLM layer representations.
Rather, it is a tractable model for isolating the high-dimensional mechanism by which ridge regularization changes the learned direction.
This mechanism predicts qualitative regularization trends, including non-monotonic dependence on $\lambda$, and is complementary to proportional-limit analyses of separable linear classification \citep{montanari2025maxmargin,deng2021doubledescent}.
It also predicts ratio-controlled behavior: at fixed representation dimension $p$, changing the number of examples $n$ should primarily affect performance through the aspect ratio $\delta=n/p$.
Our real-embedding subsampling experiment confirms this structural prediction: as $n$ varies, \textsc{Raptor}'s accuracy is largely governed by $\delta$ rather than the absolute scale of $n$.

This interpretation is consistent with a broader universality perspective in high-dimensional statistics, where proportional-limit phenomena for convex estimators can persist beyond exact Gaussian designs under suitable regularity and moment-matching conditions \citep{oymak2018universality,panahi2017universal}.
In our setting, the practical role of the theory is explanatory rather than prescriptive: to explain why varying $\lambda$ changes accuracy and directional stability, and why the resulting proportional trend remains visible on real LLM layer representations.


\subsection{Model and proportional regime}
We observe i.i.d.\ data $(x_i,y_i)_{i=1}^n$ with $x_i\in\mathbb{R}^p$ and labels $y_i\in\{\pm1\}$.
Assume the Gaussian design
\begin{equation}
 x_i \sim \mathcal{N}\!\left(0,\frac{1}{p}I_p\right),\qquad i=1,\dots,n,
\end{equation}
and a logistic teacher with parameter $\beta^\star\in\mathbb{R}^p$ satisfying $\|\beta^\star\|_2^2=p\kappa^2$ for a constant signal level $\kappa>0$:
\begin{equation}
\begin{aligned}
\Pbb(y_i=+1\mid x_i) = \sigm(x_i^\top\beta^\star),\quad
\Pbb(y_i=-1\mid x_i) = \sigm(-x_i^\top\beta^\star),\quad
\sigm(t) = \frac{1}{1+e^{-t}}.
\end{aligned}
\end{equation}

We study the proportional limit $n,p\to\infty$ with $\delta:=\lim_{p\to\infty}\frac{n}{p}\in(0,\infty).$

\subsection{Ridge logistic regression}
Let $\ell(y,t)=\log(1+e^{-yt})$ and $\rho(t)=\log(1+e^t)$.
Ridge logistic regression solves
\begin{equation}\label{eq:rlr_main}
\begin{aligned}
\hat\beta\in\argmin_{\beta\in\mathbb{R}^p}\Big\{&\ \frac{1}{n}\sum_{i=1}^n \ell\big(y_i,x_i^\top\beta\big)
+\frac{\lambda}{2p}\|\beta\|_2^2\Big\},\qquad \lambda>0.
\end{aligned}
\end{equation}

A convenient rescaling for the analysis is $z:=\beta/\sqrt{p}$, so that $x_i^\top\beta$ becomes a standard Gaussian bilinear form.
The key geometric quantities of the optimizer are its \emph{alignment} with the teacher direction and its \emph{orthogonal energy}:
\begin{equation}\label{eq:alpha_sigma_def}
\begin{aligned}
\alpha = \langle \hat z,\, v\rangle,\quad \sigma = \|\hat z-\alpha v\|_2,\quad
 v = \beta^\star/\|\beta^\star\|_2.
\end{aligned}
\end{equation}

\subsection{Fixed-point characterization}
To state the proportional-limit characterization compactly, we introduce the proximal operator of $\rho$ purely as notation (no algorithmic use is needed).
Define the proximal map of $\rho$ by
\begin{equation}\label{eq:prox_eta_def}
\begin{aligned}
\eta_{\gamma}(u) &:= \prox_{\gamma\rho}(u)
= \argmin_{t\in\mathbb{R}}\Big\{\rho(t)+\frac{1}{2\gamma}(t-u)^2\Big\},\qquad \gamma>0.
\end{aligned}
\end{equation}
Equivalently, $\eta_\gamma(u)$ is the unique solution of $\eta_\gamma(u)+\gamma\sigm\big(\eta_\gamma(u)\big)=u.$

\begin{theorem}[Ridge logistic regression in the proportional regime; adapted from \cite{salehi2019impact}]\label{thm:fixed_point}
Assume the model above with $n,p\to\infty$ and
$n/p\to\delta\in(0,\infty)$, and fix $\lambda>0$.
Suppose that the following fixed-point system has a unique solution
$(\bar\alpha,\bar\sigma,\bar\gamma)
\in\mathbb{R}\times(0,\infty)^2$.
Then the random pair $(\alpha,\sigma)$ defined above converges
in probability to $(\bar\alpha,\bar\sigma)$, where the limiting
parameters satisfy
\begin{align}
1 &=\frac{2\delta}{\bar\sigma^2}\,\E\Big[\sigm(-\kappa Z_1)\,\big(V-\eta_{\bar\gamma}(V)\big)^2\Big]
\label{eq:fp1}\\[0.35em]
\frac{\bar\alpha}{\delta}
&=-2\kappa\,\E\Big[\sigm(-\kappa Z_1)\big(1-\sigm(-\kappa Z_1)\big)\,\eta_{\bar\gamma}(V)\Big]
\label{eq:fp2}\\[0.35em]
1-\frac{1}{\delta}+\bar\gamma\lambda
&=\E\Bigg[\frac{2\,\sigm(-\kappa Z_1)}{1+\bar\gamma\,\sigm\big(\eta_{\bar\gamma}(V)\big)\big(1-\sigm\big(\eta_{\bar\gamma}(V)\big)\big)}\Bigg]
\label{eq:fp3}
\end{align}

where $Z_1,Z_2\stackrel{\mathrm{i.i.d.}}{\sim}\mathcal{N}(0,1)$ and $V=\bar\alpha Z_1+\bar\sigma Z_2$.
\end{theorem}

\subsection{Deterministic limit of test accuracy}
Let $(x,y)$ be an independent test pair from the same teacher model.
Consider the zero-threshold classifier
$\hat y(x)=\operatorname{sign}(x^\top\hat\beta)$.
Define the teacher score $U:=x^\top\beta^\star$ and the learned
score $S:=x^\top\hat\beta$.
Gaussian rotational invariance and Theorem~\ref{thm:fixed_point} yield the following
joint limit:
\begin{equation}\label{eq:test_scores_joint}
    (U,S)\xrightarrow{d}
    \bigl(\kappa Z,\bar\alpha Z+\bar\sigma W\bigr),
    \qquad
    Z,W\overset{\mathrm{i.i.d.}}{\sim}\mathcal{N}(0,1).
\end{equation}

\begin{proposition}[Asymptotic test accuracy]\label{prop:acc}
Under the conditions of Theorem~\ref{thm:fixed_point}, the out-of-sample classification accuracy converges to
\begin{equation}
\begin{split}
\mathrm{Acc}(\delta, \lambda, \kappa) = \E_Z \Big[
\ \sigm(\kappa Z)\,\PhiG\!\left(\frac{\bar\alpha}{\bar\sigma}Z\right)
+ \ \sigm(-\kappa Z)\left(1 - \PhiG\!\left(\frac{\bar\alpha}{\bar\sigma}Z\right)\right)
\Big]
\end{split}
\end{equation}
where $Z\sim\mathcal{N}(0,1)$ and $(\bar\alpha,\bar\sigma)$ are
the first two components of the unique fixed-point solution
specified in Theorem~\ref{thm:fixed_point}.
\end{proposition}

\paragraph{How $\lambda$ affects directional stability}
Our robustness metric is \emph{directional stability}, defined as the mean absolute pairwise cosine similarity between concept vectors learned across ablation runs.
The fixed-point parameters $(\bar\alpha,\bar\sigma)$ admit a simple geometric interpretation.

Defining $\hat v=\hat z/\|\hat z\|_2$, the orthogonal decomposition
in Equation~\ref{eq:alpha_sigma_def} and Theorem~\ref{thm:fixed_point} give
\begin{equation}
    \langle\hat v,v\rangle
    =
    \frac{\alpha}{\sqrt{\alpha^2+\sigma^2}}
    \xrightarrow{P}
    \frac{\bar\alpha}{\sqrt{\bar\alpha^2+\bar\sigma^2}}.
\end{equation}

Under a stylized high-dimensional approximation in which two probes trained on independent subsamples yield orthogonal noise components, the cosine similarity between the resulting directions concentrates as
\begin{equation}\label{eq:cosine_stability_approx}
\cos(\hat v^{(1)},\hat v^{(2)}) \approx \frac{\bar\alpha^2}{\bar\alpha^2+\bar\sigma^2}.
\end{equation}
Thus, increasing the signal component or suppressing orthogonal energy improves directional stability, providing a direct link between regularization and robustness.


\subsection{Validating high-dimensional structure on real embeddings}
\label{sec:exp_ab}

We test whether the proportional-theory prediction that accuracy is governed by the aspect ratio $\delta=n/p$ is visible in real LLM embeddings.
For each fixed model--dataset--layer tuple, we vary $n$ by stratified subsampling ($6$ fractions, $5$ seeds) at fixed representation dimension $p$, and compare Raptor's held-out accuracy $\mathrm{Acc}_{\text{true}}$ with a theory-inspired predictor $\mathrm{Acc}_{\text{pred}}$; details are in Appendix~\ref{app:validating}.
Across $12$ settings (2 models $\times$ 3 datasets $\times$ 2 layers), $\mathrm{Acc}_{\text{pred}}$ tracks the $\delta$ sweep well, with median Spearman $0.86$ and median Pearson $0.90$ (best over the ridge grid); full results are reported in \autoref{tab:expA_structure_validation_summary}.
This suggests that the proportional theory captures the dominant accuracy trend induced by changing $n$, despite the non-Gaussian and correlated structure of real LLM embeddings.

\section{Discussion and Conclusion}
\label{sec:discussion}

In this work, we revisited the foundations of linear probing for inference-time intervention.
While the literature often treats logistic regression as a static baseline, our findings demonstrate that its behavior in modern probing settings is heavily dependent on regularization and protocol choices.
We introduce \textsc{Raptor} to formalize this insight: by reducing the probe design to a single, validation-selected ridge parameter $\lambda$, we achieve a minimal yet highly effective standard for concept extraction.

Empirically, \textsc{Raptor} challenges the assumption that accurate concept probing requires complex estimators.
Across our benchmark, it achieves competitive accuracy while offering greater directional stability than xRFM under the evaluated ablations.
This establishes a crucial practical takeaway: a rigorously tuned ridge-logistic probe serves as a formidable reference point for evaluating more complex estimators in activation steering.

To ground these empirical successes, we complemented our benchmark with a high-dimensional theoretical analysis of ridge logistic regression.
Although our stylized Gaussian teacher--student model simplifies the complex distribution of real LLM representations, it provides analytical insight into why regularization is essential for recovering stable directions in high-dimensional spaces.
Future work can extend this theoretical framework to encompass more realistic feature dependencies, thereby bridging the gap between statistical theory and the practical dynamics of LLMs.

\paragraph{Broader impacts.}
\textsc{Raptor} is intended to support analysis and controllability of language-model representations by making concept-vector estimation more accurate, stable, and efficient.
The same capability could also be misused to manipulate model behavior or amplify undesirable concepts, so downstream deployment of steering methods should include task-specific safety evaluation and appropriate monitoring.

\begin{ack}
We thank the anonymous reviewers for their helpful feedback.
We acknowledge the computing resources and staff support of
Quest, Northwestern University's high-performance computing
facility. 

The authors declare no competing interests.
\end{ack}

\newpage
\bibliographystyle{unsrt}
\bibliography{refs}
\newpage
\appendix

\section{CGMT derivation of Theorem~\ref{thm:fixed_point}}\label{app:cgmt}
This appendix gives a CGMT-based derivation of Theorem~\ref{thm:fixed_point}.
After matching the signal-alignment normalization, the resulting fixed-point equations are equivalent to those in \cite{salehi2019impact}.
The asymptotic optimizer-convergence step uses that result, with its uniqueness assumption stated explicitly below.
The main text states the final fixed-point system (Theorem~\ref{thm:fixed_point}) and the limiting test accuracy (Proposition~\ref{prop:acc});
here we give a CGMT-based route from the ERM \eqref{eq:rlr_main} to the deterministic characterization.
\providecommand{\R}{\mathbb{R}}
\providecommand{\Pbb}{\mathbb{P}}
\providecommand{\norm}[1]{\left\lVert#1\right\rVert}
\providecommand{\ip}[2]{\left\langle #1,#2\right\rangle}
\providecommand{\sigm}{\sigma}
\providecommand{\rhoL}{\rho}
\providecommand{\rhoc}{\rho^{\!*}}
\providecommand{\PhiG}{\Phi}

\providecommand{\bbeta}{\bm{\beta}}
\providecommand{\btheta}{\bm{\theta}}
\providecommand{\bSigma}{\bm{\Sigma}}
\providecommand{\bmu}{\bm{\mu}}
\providecommand{\bnu}{\bm{\nu}}

\providecommand{\MatM}{\mathbf{M}}
\providecommand{\MatA}{\mathbf{A}}
\providecommand{\vecz}{\mathbf{z}}
\providecommand{\vecu}{\mathbf{u}}
\providecommand{\vecs}{\mathbf{s}}
\providecommand{\vect}{\mathbf{t}}
\providecommand{\vecw}{\mathbf{w}}
\providecommand{\vecy}{\mathbf{y}}
\providecommand{\vecg}{\mathbf{g}}
\providecommand{\vech}{\mathbf{h}}

\subsection{Model, estimator, and scaling}

\subsubsection{Data model and proportional regime}
We observe a feature matrix and labels
\[
\mathbf{X}:=\begin{bmatrix}\mathbf{x}_1^\top\\ \vdots\\ \mathbf{x}_n^\top\end{bmatrix}\in\R^{n\times p},
\qquad
\vecy:=(y_1,\dots,y_n)\in\{\pm1\}^n.
\]
We work in the \emph{proportional} asymptotic regime
\[
\delta\in(0,\infty),
\qquad
n,p\to\infty\ \text{with }\frac{n}{p}\to\delta.
\]
The three scalar parameters that will index the final accuracy are:
\begin{itemize}[leftmargin=2em,itemsep=0.2em]
\item $\delta=\lim n/p$: limiting sample-to-dimension ratio,
\item $\lambda>0$: ridge strength,
\item $\kappa>0$: signal strength of the teacher.
\end{itemize}

Assume
\[
\mathbf{X}=\frac{1}{\sqrt{p}}\MatM,\qquad \MatM\in\R^{n\times p}\ \text{has i.i.d. }\mathcal N(0,1)\ \text{entries}.
\]
Equivalently, the rows satisfy $\mathbf{x}_i\sim\mathcal N(0,\mathbf I_p/p)$ independently.

Fix a \emph{true parameter} $\bbeta^\star\in\R^p$ with
\[
\|\bbeta^\star\|^2 = p\kappa^2,\qquad \kappa>0\ \text{constant}.
\]
Let $\mathbf v:=\bbeta^\star/\|\bbeta^\star\|$ so that $\|\mathbf v\|=1$ and $\bbeta^\star=\kappa\sqrt p\,\mathbf v$.
Define the teacher score for sample $i$:
\[
U_i:=\mathbf{x}_i^\top\bbeta^\star
=\frac{1}{\sqrt p}\MatM_{i:}\bbeta^\star
=\kappa\,(\MatM\mathbf v)_i.
\]
Conditioned on $\mathbf X$ (equivalently on $\MatM$), the labels are independent with
\[
\Pbb(y_i=+1\mid \mathbf X)=\sigm(U_i),
\qquad
\Pbb(y_i=-1\mid \mathbf X)=\sigm(-U_i),
\]
where $\sigm(u):=1/(1+e^{-u})$.

\subsubsection{Estimator (ridge logistic regression)}
Define the logistic loss $\ell(y,t):=\log(1+e^{-yt})=\rhoL(-yt)$ and the log-partition function
\[
\rhoL(u):=\log(1+e^{u}).
\]
We consider the ridge-logistic scaling used in \cite{salehi2019impact}:
\begin{equation}\label{eq:ERM-beta}
\begin{aligned}
\hat\bbeta\in\argmin_{\bbeta\in\R^p}\Big\{&\ \frac{1}{n}\sum_{i=1}^n \ell\big(y_i,\mathbf{x}_i^\top\bbeta\big)
+\frac{\lambda}{2p}\|\bbeta\|^2\Big\},\qquad \lambda>0.
\end{aligned}
\end{equation}
Multiplying the objective by $n$ (which does not change the minimizer) gives the equivalent form
\[
\hat\bbeta\in\argmin_{\bbeta\in\R^p}
\Big\{\ \sum_{i=1}^n \ell\big(y_i,\mathbf{x}_i^\top\bbeta\big)+\frac{n\lambda}{2p}\|\bbeta\|^2\ \Big\}.
\]

\subsubsection{Convenient rescaling}
Introduce the scaled variable
\[
\vecz := \frac{\bbeta}{\sqrt{p}}\in\R^p \quad\Longleftrightarrow\quad \bbeta=\sqrt{p}\,\vecz.
\]
Then $\mathbf{x}_i^\top\bbeta=(\MatM_{i:}\vecz)$ and $\|\bbeta\|^2=p\|\vecz\|^2$. Substituting into the unnormalized objective gives
\begin{equation}\label{eq:ERM-z}
\begin{aligned}
\hat{\vecz}\in\argmin_{\vecz\in\R^p}\Big\{&\ \sum_{i=1}^n \rhoL\!\big(-y_i(\MatM_{i:}\vecz)\big)
+ \frac{n\lambda}{2}\|\vecz\|^2\Big\}.
\end{aligned}
\end{equation}
We will characterize the limiting signal-aligned and orthogonal components of $\hat{\vecz}$, then translate back to $\hat\bbeta=\sqrt{p}\,\hat{\vecz}$.

\subsection{From ERM to a convex min--max (Primary Optimization)}

\begin{lemma}[Convex conjugate of $\rhoL(u)=\log(1+e^u)$]\label{lem:rho_conj}
Let $\rhoL(u)=\log(1+e^{u})$. Its Fenchel conjugate $\rhoc$ is
\[
\rhoc(s)=
\begin{cases}
s\log s + (1-s)\log(1-s), & s\in(0,1),\\
0, & s\in\{0,1\},\\
+\infty, & \text{otherwise}.
\end{cases}
\]
Moreover, for all $u\in\R$,
\begin{equation}\label{eq:fenchel_rho}
\rhoL(u)=\max_{s\in[0,1]}\{\, s\,u - \rhoc(s)\,\}.
\end{equation}
The maximizer is $s^\star=\sigm(u)$.
\end{lemma}

Starting from \eqref{eq:ERM-z}, apply Lemma~\ref{lem:rho_conj} coordinate-wise with $u=-y_i(\MatM_{i:}\vecz)$:
\[
\rhoL\big(-y_i\MatM_{i:}\vecz\big)
=\max_{w_i\in[0,1]}\Big\{-w_i\,y_i\,(\MatM_{i:}\vecz)-\rhoc(w_i)\Big\}.
\]
Let $\vecw=(w_1,\dots,w_n)\in[0,1]^n$. Plugging into \eqref{eq:ERM-z} and exchanging the finite sum with the maximization gives the saddle formulation
\begin{equation}\label{eq:PO_raw}
\boxed{
\Phi(\MatM,\vecy)
:=
\min_{\vecz\in\mathbb{R}^p}\ \max_{\vecw\in[0,1]^n}
\Big\{
 -\vecw^{\top}\diag(\vecy)\MatM\vecz 
 - \sum_{i=1}^n \rhoc(w_i)
+ \frac{n\lambda}{2}\|\vecz\|^2
\Big  \}
}
\end{equation}

The vector $\vecw\in[0,1]^n$ is the Fenchel dual variable arising from \eqref{eq:fenchel_rho}; there is no additional sign constraint tied to $\vecy$.

\paragraph{Why an independence reduction is necessary.}
The Gaussian matrix $\MatM$ and the labels $\vecy$ are dependent: $\vecy$ is generated from the teacher score $\MatM\mathbf v$.
To apply CGMT, we isolate \emph{all} dependence into a single Gaussian vector and keep an independent Gaussian matrix in the remaining bilinear term.

\subsection{Isolating label dependence via orthogonal decomposition}

\subsubsection{Gaussian object that generates $\vecy$}
Define the signal-direction score vector
\[
\vecu := \MatM\mathbf v\in\R^n.
\]
Since $\MatM$ has i.i.d.\ $\mathcal N(0,1)$ entries and $\|\mathbf v\|=1$, we have $\vecu\sim\mathcal N(0,\mathbf I_n)$. Moreover,
\[
\Pbb(y_i=+1\mid u_i)=\sigm(\kappa u_i),
\qquad
\Pbb(y_i=-1\mid u_i)=\sigm(-\kappa u_i),
\]
so the entire dependence of $\vecy$ on $\MatM$ is through the coordinates $\{u_i\}_{i=1}^n$.

\begin{lemma}[Gaussian orthogonal decomposition]\label{lem:gauss_decomp}
Let $\MatM\in\R^{n\times p}$ have i.i.d.\ $\mathcal N(0,1)$ entries and let $\mathbf v\in\R^p$ be deterministic with $\|\mathbf v\|=1$. Set $\vecu:=\MatM\mathbf v$. Then there exists a random matrix $\MatM_\perp$ such that
\[
\MatM=\vecu\,\mathbf v^\top+\MatM_\perp,\qquad \MatM_\perp\mathbf v=\mathbf 0,
\]
and $\MatM_\perp$ is independent of $\vecu$. In an orthonormal basis of $\mathbf v^\perp$, the coordinates of $\MatM_\perp$ are i.i.d.\ $\mathcal N(0,1)$.
\end{lemma}

\subsubsection{Decompose both the matrix and the primal variable}
Using Lemma~\ref{lem:gauss_decomp}, write
\begin{equation}\label{eq:M_decomp}
\MatM = \vecu\,\mathbf v^{\top} + \MatM_{\perp},
\qquad \MatM_{\perp}\mathbf v=\mathbf 0.
\end{equation}
Decompose the optimization variable similarly:
\begin{equation}\label{eq:z_decomp}
\vecz = \alpha\,\mathbf v + \vecz_\perp,
\qquad \alpha\in\R,\ \vecz_\perp\in\mathbf v^\perp.
\end{equation}
Then $\|\vecz\|^2=\alpha^2+\|\vecz_\perp\|^2$ and
\[
\MatM\vecz
= (\vecu\,\mathbf v^\top)\vecz+\MatM_\perp\vecz_\perp
=\alpha\,\vecu+\MatM_\perp\vecz_\perp.
\]
Plugging \eqref{eq:M_decomp}--\eqref{eq:z_decomp} into \eqref{eq:PO_raw} gives
\begin{equation}\label{eq:PO_decomp}
\begin{aligned}
\Phi
&=\min_{\alpha\in\R,\ \vecz_\perp\in\mathbf v^\perp}
  \max_{\vecw\in[0,1]^n}
\Big\{
-\alpha\,\vecw^\top\diag(\vecy)\vecu
-\vecw^\top\diag(\vecy)\MatM_\perp\vecz_\perp \\
&\hspace{4em}
-\sum_{i=1}^n\rhoc(w_i)
+\frac{n\lambda}{2}\big(\alpha^2+\|\vecz_\perp\|^2\big)
\Big\}.
\end{aligned}
\end{equation}
Define the two objects that remain throughout the CGMT step:
\[
\vecs:=\diag(\vecy)\vecu\in\R^n,
\qquad
\MatA:=\left.\diag(\vecy)\MatM_\perp\right|_{\mathbf v^\perp}.
\]
We represent the second operator in a fixed orthonormal basis of $\mathbf v^\perp$, so that $\MatA\in\R^{n\times(p-1)}$.
When multiplying by $\MatA$, we identify $\vecz_\perp$ with its coordinates in this basis.
Then
\begin{equation}\label{eq:PO_for_CGMT}
\begin{aligned}
\Phi
&=\min_{\alpha\in\R,\ \vecz_\perp\in\mathbf v^\perp}
  \max_{\vecw\in[0,1]^n}
\Big\{
-\alpha\,\vecw^\top\vecs
-\vecw^\top\MatA\vecz_\perp \\
&\hspace{4em}
-\sum_{i=1}^n\rhoc(w_i)
+\frac{n\lambda}{2}\big(\alpha^2+\|\vecz_\perp\|^2\big)
\Big\}.
\end{aligned}
\end{equation}

Conditional on $(\vecu,\vecy)$, and hence also on $\vecs$, the entries of $\MatA$ in this basis are i.i.d.\ standard Gaussian.
Its conditional distribution does not depend on $(\vecu,\vecy)$, so $\MatA$ is independent of $\vecs$.
Indeed, $\MatM_\perp$ is independent of $(\vecu,\vecy)$, since the labels are generated from $\vecu$ alone, and multiplying its rows by $\pm1$ preserves its Gaussian distribution.

\subsection{CGMT reduction: auxiliary optimization}

\subsubsection{CGMT form used}
We invoke the convex Gaussian min--max theorem (CGMT) in the standard bilinear form; see, e.g., \cite{thrampoulidis2015regularized,thrampoulidis2018mestimators}.

\begin{theorem}[CGMT, specialized form]\label{thm:cgmt_spec}
Let $\MatA\in\R^{n\times(p-1)}$, $\vecg\in\R^n$, and $\vech\in\R^{p-1}$ have mutually independent standard Gaussian entries.
Let $\mathcal S\subset\R\times\R^{p-1}$ and $\mathcal T\subset\R^n$ be nonempty compact convex sets, and let $\Psi(\alpha,\vecz_\perp,\vecw)$ be deterministic and continuous, convex in $(\alpha,\vecz_\perp)$, and concave in $\vecw$.
Define
\begin{align*}
\Phi_{\mathrm{PO}}
&:=\min_{(\alpha,\vecz_\perp)\in\mathcal S}
   \max_{\vecw\in\mathcal T}
\big\{\vecw^\top\MatA\vecz_\perp
+\Psi(\alpha,\vecz_\perp,\vecw)\big\},\\
\Phi_{\mathrm{AO}}
&:=\min_{(\alpha,\vecz_\perp)\in\mathcal S}
   \max_{\vecw\in\mathcal T}
\big\{\|\vecz_\perp\|\,\vecg^\top\vecw
+\|\vecw\|\,\vech^\top\vecz_\perp
+\Psi(\alpha,\vecz_\perp,\vecw)\big\}.
\end{align*}
For every $\mu\in\R$ and $t>0$,
\[
\Pbb\big(|\Phi_{\mathrm{PO}}-\mu|>t\big)
\le 2\Pbb\big(|\Phi_{\mathrm{AO}}-\mu|\ge t\big).
\]
The usual strict-gap comparison for a restricted minimizing set transfers concentration of optimizer summaries from the AO to the PO.
\end{theorem}

Here $\alpha$ is an auxiliary minimizing coordinate that enters only through $\Psi$.
This form follows from the standard CGMT by introducing a positive scale $\varepsilon$, appending an independent Gaussian column acting on $\varepsilon\alpha$, and letting $\varepsilon\downarrow0$.
Compactness makes both optimization values continuous under this limit.

\subsubsection{Apply CGMT to \eqref{eq:PO_for_CGMT}}
In \eqref{eq:PO_for_CGMT}, conditional on $\vecs$, the only random Gaussian object is $\MatA$ and it appears only in the bilinear term $-\vecw^\top\MatA\vecz_\perp$.
The compactness requirement does not change the optimizer: the PO objective is at least $n\lambda\|\vecz\|^2/2$, while its value at $\vecz=\mathbf0$ is $n\log2$.
Thus the primal domain may be restricted to a closed ball of any fixed radius greater than $\sqrt{2\log2/\lambda}$.
The same bound holds for the AO below: choosing $\vecw=\mathbf0$ gives the quadratic lower bound, and the value at $(\alpha,\vecz_\perp)=(0,\mathbf0)$ is $n\log2$.
We omit this inactive ball constraint in the following displays.

Absorb the minus sign by redefining $\MatA\leftarrow-\MatA$ (distribution unchanged). CGMT then yields the auxiliary optimization
\begin{equation}\label{eq:AO_0}
\begin{aligned}
\phi&=
\min_{\alpha\in\R,\ \vecz_\perp\in\mathbf v^\perp}
\max_{\vecw\in[0,1]^n}
\Big\{\ 
-\alpha\,\vecw^\top\vecs
+\|\vecz_\perp\|\,\vecg^\top\vecw
 +\|\vecw\|\,\vech^\top\vecz_\perp \\
&\hspace{4em}-\sum_{i=1}^n\rhoc(w_i)
+\frac{n\lambda}{2}\big(\alpha^2+\|\vecz_\perp\|^2\big)
\Big\}.
\end{aligned}
\end{equation}
Here $\vecg\in\R^n$ and $\vech\in\R^{p-1}$ are independent standard Gaussian vectors, independent of $(\vecu,\vecy)$.

\subsection{Scalarization and decoupling of the auxiliary optimization}

\subsubsection{\texorpdfstring{Eliminate the direction of $\vecz_\perp$}{Eliminate the direction of z_perp}}

Fix $\alpha$, $\vecw$, and let $r:=\|\vecz_\perp\|\ge0$. The only term depending on the direction of $\vecz_\perp$ is $\|\vecw\|\,\vech^\top\vecz_\perp$.
Minimizing over all $\vecz_\perp\in\mathbf v^\perp$ with $\|\vecz_\perp\|=r$ yields
\[
\min_{\|\vecz_\perp\|=r}\ \|\vecw\|\,\vech^\top\vecz_\perp
=-\|\vecw\|\,r\,\|\vech\|.
\]
For $\vech\ne\mathbf0$, the minimizing direction is $-\vech/\|\vech\|$, independently of $\vecw$; if $\vech=\mathbf0$, every direction gives the same value.
Thus the same direction minimizes the expression for every dual vector, and \eqref{eq:AO_0} becomes
\begin{equation}\label{eq:AO_1}
\begin{aligned}
\phi=\min_{\alpha\in\R,\ r\ge 0}\ \max_{\vecw\in[0,1]^n}\Big\{
-\alpha\,\vecw^\top \vecs
+r\,\vecg^\top \vecw
-r\,\|\vech\|\,\|\vecw\|
-\sum_{i=1}^n\rhoc(w_i)
+\frac{n\lambda}{2}(\alpha^2+r^2)
\Big\}.
\end{aligned}
\end{equation}

\subsubsection{Two variational identities used to decouple norms and squares}
\begin{lemma}[Linearize a norm]\label{lem:norm_variational}
For any $a\ge 0$ and any vector $\vecw\in\R^n$,
\[
-a\|\vecw\|
=\sup_{\tau>0}\left\{-\frac{\tau}{2}\|\vecw\|^2-\frac{a^2}{2\tau}\right\}.
\]
\end{lemma}

\begin{lemma}[Linearize a negative square]\label{lem:square_variational}
For any $c>0$ and scalar $u\in\R$,
\[
-\frac{u^2}{2c}=\min_{\gamma\in\R}\left\{\frac{c}{2}\gamma^2-\gamma u\right\}.
\]
\end{lemma}

\subsubsection{Decouple $-\|\vecw\|$}
Apply Lemma~\ref{lem:norm_variational} to the coupling term with $a=r\|\vech\|$:
\[
-r\|\vech\|\,\|\vecw\|
=\sup_{\tau>0}\left\{-\frac{\tau}{2}\|\vecw\|^2-\frac{r^2\|\vech\|^2}{2\tau}\right\}.
\]
Substituting into \eqref{eq:AO_1} gives
\begin{equation}\label{eq:AO_2}
\begin{aligned}
\phi&=\min_{\alpha\in\R,\ r\ge 0}\ \sup_{\tau>0}\ \max_{\vecw\in[0,1]^n}
\Big\{
-\alpha\,\vecw^\top \vecs
+r\,\vecg^\top \vecw
-\frac{\tau}{2}\|\vecw\|^2 \\
&\hspace{4em}-\sum_{i=1}^n\rhoc(w_i)
+\frac{n\lambda}{2}(\alpha^2+r^2)
-\frac{r^2\|\vech\|^2}{2\tau}
\Big\}.
\end{aligned}
\end{equation}

\subsubsection{Eliminate $\alpha$ explicitly}
For fixed $r$, the objective in \eqref{eq:AO_1} is convex and uniformly coercive in $\alpha$, and concave in $\vecw\in[0,1]^n$.
The convex--concave minimax equality therefore allows $\min_\alpha$ and $\max_{\vecw}$ to be exchanged in \eqref{eq:AO_1}, before introducing $\tau$.
For fixed $(r,\vecw)$, the $\alpha$-dependent part is
\[
-\alpha\,\vecw^\top\vecs+\frac{n\lambda}{2}\alpha^2,
\]
whose minimum occurs at $\alpha^\star=(\vecw^\top\vecs)/(n\lambda)$ and equals $-(\vecw^\top\vecs)^2/(2n\lambda)$.
Applying the norm identity after this elimination, and commuting the two suprema, gives
\begin{equation}\label{eq:AO_3}
\begin{aligned}
\phi&=\min_{r\ge 0}\ \sup_{\tau>0}\ \max_{\vecw\in[0,1]^n}
\Big\{r\,\vecg^\top \vecw
-\frac{\tau}{2}\|\vecw\|^2
-\sum_{i=1}^n\rhoc(w_i)
+\frac{n\lambda}{2}r^2
-\frac{r^2\|\vech\|^2}{2\tau}
-\frac{(\vecw^\top\vecs)^2}{2n\lambda}
\Big\}.
\end{aligned}
\end{equation}

\subsubsection{Decouple $(\vecw^\top\vecs)^2$}
Apply Lemma~\ref{lem:square_variational} with $u=\vecw^\top\vecs$ and $c=n\lambda$, reintroducing the eliminated scalar $\alpha$:
\[
-\frac{(\vecw^\top\vecs)^2}{2n\lambda}
=\min_{\alpha\in\R}\left\{\frac{n\lambda}{2}\alpha^2-\alpha\,\vecw^\top\vecs\right\}.
\]
For each fixed $(r,\tau)$, the resulting objective is convex and uniformly coercive in $\alpha$ and concave in the compact dual variable $\vecw$.
The convex--concave minimax equality gives
\begin{equation}\label{eq:AO_4}
\begin{aligned}
\phi
&=\min_{r\ge0}\ \sup_{\tau>0}\ \min_{\alpha\in\R}\ \max_{\vecw\in[0,1]^n}
\Big\{
\sum_{i=1}^n\Big[(r g_i-\alpha s_i)w_i
-\frac{\tau}{2}w_i^2-\rhoc(w_i)\Big]\\
&\hspace{5em}
+\frac{n\lambda}{2}r^2
-\frac{r^2\|\vech\|^2}{2\tau}
+\frac{n\lambda}{2}\alpha^2
\Big\}.
\end{aligned}
\end{equation}
Now the maximization over $\vecw$ is coordinate-wise separable.

\subsection{Coordinate-wise maximization and the logistic proximal map}

\begin{lemma}[Prox definition]\label{lem:prox_def}
For a proper closed convex $f:\R\to(-\infty,+\infty]$ and $\gamma>0$,
\[
\prox_{\gamma f}(v):=\argmin_{u\in\R}\Big\{ f(u)+\frac{1}{2\gamma}(u-v)^2\Big\}
\]
\end{lemma}

Fix $(r,\tau,\alpha)$ and define $a_i:=r g_i-\alpha s_i$.
For each coordinate we need
\[
\max_{w\in[0,1]}\Big\{ a_i w - \frac{\tau}{2}w^2 - \rhoc(w)\Big\}
\]
Completing the square gives
\[\begin{aligned}
&\max_{w\in[0,1]}\Big\{ a_i w - \frac{\tau}{2}w^2 - \rhoc(w)\Big\}
=\frac{a_i^2}{2\tau}\\
&-\min_{w\in[0,1]}\Big\{\rhoc(w)+\frac{\tau}{2}\Big(w-\frac{a_i}{\tau}\Big)^2\Big\}
\end{aligned}\]
By Lemma~\ref{lem:prox_def}, the maximizer is
\[
w_i^\star=\prox_{(1/\tau)\rhoc}\!\left(\frac{a_i}{\tau}\right)
\]

\begin{lemma}[Moreau decomposition (scalar form)]\label{lem:moreau}
Let $f$ be proper closed convex with conjugate $f^\star$. For any $\gamma>0$ and $v\in\R$,
\[
\prox_{\gamma f}(v) + \gamma\,\prox_{f^\star/\gamma}\!\left(\frac{v}{\gamma}\right)=v
\]
Equivalently,
\[
\prox_{(1/\gamma) f^\star}(v)=v-\frac{1}{\gamma}\,\prox_{\gamma f}(\gamma v)
\]
\end{lemma}

\subsubsection{Convert $\prox$ of $\rhoc$ into $\prox$ of $\rhoL$}
Using Lemma~\ref{lem:moreau} with $f=\rhoL$ and $f^\star=\rhoc$ and $\gamma=\tau$,
\[
\prox_{(1/\tau)\rhoc}(v)=v-\frac{1}{\tau}\,\prox_{\tau\rhoL}(\tau v)
\]
Apply this with $v=a_i/\tau$ and define
\[
\eta_i:=\prox_{\tau\rhoL}(a_i)
\]
Then
\begin{equation}\label{eq:w_star_eta}
w_i^\star
=\frac{a_i}{\tau}-\frac{1}{\tau}\,\eta_i
\end{equation}

\subsubsection{Implicit equation for the logistic proximal map}
Since $\rhoL'(t)=\sigm(t)$, the first-order condition for $\eta=\prox_{\tau\rhoL}(a)$ is
\[
0=\rhoL'(\eta)+\frac{1}{\tau}(\eta-a)
\quad\Longleftrightarrow\quad
\boxed{\ \eta+\tau\sigm(\eta)=a\ }.
\]
This has a unique solution for each $(a,\tau)$ because the left-hand side is strictly increasing in $\eta$ and has range $\R$.

Define the Moreau envelope of $\rhoL$ by
\[
M_{\rhoL}(a;\tau)
:=\min_{u\in\R}\left\{\rhoL(u)+\frac{(u-a)^2}{2\tau}\right\}
=\max_{w\in[0,1]}\left\{a w-\frac{\tau}{2}w^2-\rhoc(w)\right\}.
\]
Thus the coordinate-wise optimal value is $M_{\rhoL}(a_i;\tau)$.

\subsection{Deterministic fixed-point characterization}

Write the estimator in the teacher geometry:
\[
\hat{\vecz}=\alpha\,\mathbf v+\vecz_\perp,
\qquad \alpha=\ip{\hat{\vecz}}{\mathbf v},
\qquad \sigma:=\|\vecz_\perp\|.
\]
Use $\sigma$ for the radius $r$ and $\gamma$ for the proximal parameter $\tau$ in \eqref{eq:AO_2}.
Substituting the coordinate-wise optimal values gives
\[
\frac{\phi}{n}
=\min_{\alpha\in\R,\ \sigma\ge0}\ \sup_{\gamma>0}
\left\{
\frac{\lambda}{2}(\alpha^2+\sigma^2)
-\frac{\sigma^2\|\vech\|^2}{2n\gamma}
+\frac1n\sum_{i=1}^n M_{\rhoL}(-\alpha s_i+\sigma g_i;\gamma)
\right\}.
\]
Let $Z_1,Z_2$ be independent standard Gaussian variables and define
\[
V:=\alpha Z_1+\sigma Z_2,
\qquad q:=\rhoL'(-\kappa Z_1).
\]
For any integrable function $f$, Gaussian symmetry and the label model imply
\[
\E\big[f(-\alpha y_i u_i+\sigma g_i)\big]
=2\E\big[\rhoL'(-\kappa Z_1)f(\alpha Z_1+\sigma Z_2)\big].
\]
Since $\|\vech\|^2/n\to1/\delta$, the pointwise population limit of the scalar objective is
\begin{equation}\label{eq:cgmt_population_objective}
F(\alpha,\sigma,\gamma)
:=\frac{\lambda}{2}(\alpha^2+\sigma^2)
-\frac{\sigma^2}{2\delta\gamma}
+2\E\big[q\,M_{\rhoL}(V;\gamma)\big].
\end{equation}

For $\gamma>0$, define
\[
\eta_\gamma(t):=\prox_{\gamma\rhoL}(t),
\qquad
w_\gamma(t):=\rhoL'\big(\eta_\gamma(t)\big)
=\frac{t-\eta_\gamma(t)}{\gamma}.
\]
The proximal equation and its derivative give
\[
\eta_\gamma(t)+\gamma\sigm\big(\eta_\gamma(t)\big)=t,
\qquad
w_\gamma'(t)
=\frac{\rhoL''(\eta_\gamma(t))}{1+\gamma\rhoL''(\eta_\gamma(t))}.
\]
At an interior saddle point, differentiating \eqref{eq:cgmt_population_objective} yields
\begin{align*}
0&=\lambda\alpha+2\E\big[qZ_1w_\gamma(V)\big],\\
0&=\lambda\sigma-\frac{\sigma}{\delta\gamma}
  +2\E\big[qZ_2w_\gamma(V)\big],\\
0&=\frac{\sigma^2}{2\delta\gamma^2}
  -\E\big[q\,w_\gamma(V)^2\big].
\end{align*}
Gaussian integration by parts in $Z_2$, followed by division by $\sigma>0$, gives
\[
\lambda+2\E\big[q\,w_\gamma'(V)\big]=\frac{1}{\delta\gamma}.
\]
Integration by parts in $Z_1$ then gives
\[
\frac{\alpha}{\delta\gamma}
=2\kappa\E\big[\rhoL''(-\kappa Z_1)w_\gamma(V)\big].
\]
Because $\rhoL''$ is even, $\E[\rhoL''(-\kappa Z_1)V]=0$.
Using $\gamma w_\gamma(V)=V-\eta_\gamma(V)$ and $2\E[q]=1$, these three stationarity equations reduce to \eqref{eq:R1}--\eqref{eq:R3} below.

\begin{theorem}[Fixed-point system for ridge logistic regression]\label{thm:fixed_point_app}
Assume the Gaussian teacher--student logistic model specified above, with $n,p\to\infty$, $n/p\to\delta\in(0,\infty)$, $\kappa>0$, and fixed $\lambda>0$.
Assume that \eqref{eq:R1}--\eqref{eq:R3} have a unique solution with $\bar\alpha\in\R$, $\bar\sigma>0$, and $\bar\gamma>0$.
Then, for the estimator \eqref{eq:ERM-beta},
\[
(\alpha,\sigma)
=\left(\ip{\hat{\vecz}}{\mathbf v},\;
\big\|\hat{\vecz}-\ip{\hat{\vecz}}{\mathbf v}\mathbf v\big\|\right)
\xrightarrow{\mathrm{P}}(\bar\alpha,\bar\sigma),
\]
where the limiting parameters satisfy
\begin{align}
1
&=\frac{2\delta}{\bar\sigma^2}\,
\E\Big[
\rhoL'(-\kappa Z_1)\big(V-\eta_{\bar\gamma}(V)\big)^2
\Big],
\label{eq:R1}\\[0.5em]
\frac{\bar\alpha}{\delta}
&=-2\kappa\,\E\Big[
\rhoL''(-\kappa Z_1)\eta_{\bar\gamma}(V)
\Big],
\label{eq:R2}\\[0.5em]
1-\frac{1}{\delta}+\bar\gamma\lambda
&=\E\Bigg[
\frac{2\rhoL'(-\kappa Z_1)}{1+\bar\gamma\rhoL''(\eta_{\bar\gamma}(V))}
\Bigg],
\label{eq:R3}
\end{align}
with $V=\bar\alpha Z_1+\bar\sigma Z_2$, $\rhoL'(t)=\sigm(t)$, and $\rhoL''(t)=\sigm(t)(1-\sigm(t))$.
\end{theorem}

\begin{proof}
The calculation above derives the fixed-point equations from the limiting scalar objective.
For optimizer convergence, we invoke the ridge specialization in \cite[Theorem~2]{salehi2019impact}, retaining its uniqueness assumption.
The signal multiplier in that result equals $\bar\alpha/\kappa$ in our notation; $\bar\sigma$, $\bar\gamma$, and $\lambda$ have the same normalization.
The deterministic teacher direction causes no additional restriction: Gaussian design and the ridge penalty are rotationally invariant, so the teacher may be rotated to a vector with every coordinate equal to $\kappa$.
Those constant coordinates satisfy the i.i.d.\ teacher-coordinate assumption using a point-mass distribution.
The cited convergence result therefore gives the stated limits.
\end{proof}

\subsection{Asymptotic test accuracy and its dependence on $(\delta,\lambda,\kappa)$}

\subsubsection{Joint Gaussian limit of test scores}
Draw an independent test point $\mathbf{x}\sim\mathcal N(0,\mathbf I_p/p)$ and a test label $y\in\{\pm1\}$ generated by the same teacher:
\[
\Pbb(y=+1\mid\mathbf{x})=\sigm(\mathbf{x}^\top\bbeta^\star),
\qquad
\Pbb(y=-1\mid\mathbf{x})=\sigm(-\mathbf{x}^\top\bbeta^\star).
\]
Define the teacher score and learned score by
\[
U:=\mathbf{x}^\top\bbeta^\star,
\qquad S:=\mathbf{x}^\top\hat\bbeta.
\]
Using $\hat\bbeta=\sqrt p\,\hat{\vecz}$, $\bbeta^\star=\kappa\sqrt p\,\mathbf v$, and the orthogonal decomposition of $\hat{\vecz}$, Gaussian rotational invariance gives the exact conditional distributional identity
\[
(U,S)\mid(\mathbf X,\vecy)
\ \overset{\mathrm d}{=}\ 
(\kappa Z,\alpha Z+\sigma W),
\]
where $Z,W$ are independent standard Gaussian variables, independent of the training data, and $(\alpha,\sigma)$ are fixed by those data.
Consequently, Theorem~\ref{thm:fixed_point_app} gives the joint limit
\[
(U,S)\xrightarrow{\mathrm d}
(\kappa Z,\bar\alpha Z+\bar\sigma W).
\]

\subsubsection{Closed form for accuracy of the zero-threshold classifier}
Consider the classifier $\hat y=\mathrm{sign}(S)$; ties have probability zero in the limiting model since $\bar\sigma>0$.
Let $\mathrm{Acc}$ denote the deterministic limit, in probability, of the conditional test accuracy $\Pbb(\hat y=y\mid\mathbf X,\vecy)$.
Conditional on $Z$, the test label satisfies $\Pbb(y=+1\mid Z)=\sigm(\kappa Z)$ and is independent of $W$.
The limiting learned score has distribution $\mathcal N(\bar\alpha Z,\bar\sigma^2)$ given $Z$.
Hence
\begin{equation}\label{eq:Acc_final}
\mathrm{Acc}
=\E_Z\Bigg[
\sigm(\kappa Z)\,\PhiG\!\left(\frac{\bar\alpha Z}{\bar\sigma}\right)
+\sigm(-\kappa Z)\left(1-\PhiG\!\left(\frac{\bar\alpha Z}{\bar\sigma}\right)\right)
\Bigg],
\end{equation}
where $\PhiG$ is the standard normal cumulative distribution function.
\begin{corollary}[Final performance characterization]\label{cor:final}
Under the assumptions of Theorem~\ref{thm:fixed_point_app}, the asymptotic test accuracy of ridge logistic regression is given by \eqref{eq:Acc_final}.
\end{corollary}

\subsubsection{What are the independent variables, and how does $\mathrm{Acc}$ vary?}
The accuracy is a deterministic function
\[
\mathrm{Acc}=\mathrm{Acc}(\delta,\lambda,\kappa),
\]
through the fixed-point solution $(\bar\alpha,\bar\sigma)$ in Theorem~\ref{thm:fixed_point}.
It is useful to summarize the dependence through the single \emph{effective signal-to-noise ratio}
\begin{equation}\label{eq:snr_def}
m:=\frac{\bar\alpha}{\bar\sigma}.
\end{equation}
Indeed, in \eqref{eq:Acc_final} the learned classifier enters only through $m$ (and the teacher enters through $\kappa$).

\paragraph{Interpretation via an effective margin $m$.}
For fixed $\kappa$, the accuracy expression \eqref{eq:Acc_final} is increasing in an \emph{effective margin} $m$ on the relevant branch $\bar\alpha\ge 0$: larger $m$ means the test score $S$ is more aligned with the teacher score $U$ and has less orthogonal noise relative to its signal component.
Two useful limits are:
\begin{itemize}[leftmargin=2em,itemsep=0.2em]
\item If $m=0$, then $S$ is independent of $y$ and $\mathrm{Acc}=1/2$.
\item As $m\to\infty$ at fixed $\kappa$, $\Pbb(\mathrm{sign}(S)=\mathrm{sign}(U))\to1$ and $\mathrm{Acc}\to\E_Z\big[\sigm(\kappa|Z|)\big]$
(the Bayes-optimal accuracy under the logistic teacher).
\end{itemize}

\section{Experimental details}\label{app:exp}
\subsection{Raptor algorithm details}
Details are shown in Algorithm~\ref{alg:Raptor_impl}.
\begin{algorithm}[!htb]
\caption{Raptor: train-only standardization, automatic $\lambda$ tuning, and fold-back}
\label{alg:Raptor_impl}
\begin{algorithmic}
\REQUIRE Layer-wise embeddings $\{H^{(\ell)}\}_{\ell=1}^L$, labels $y\in\{0,1\}^n$;
split indices $I_{\text{tr}}, I_{\text{val}}, I_{\text{te}}$.
\ENSURE Per-layer concept vectors $\{(\omega^{(\ell)}, b_{\text{orig}}^{(\ell)})\}_{\ell=1}^L$,
pre-refit validation accuracies $a_{\mathrm{val}}^{(\ell)}$ when available, and test accuracies.

\FOR{$\ell \leftarrow 1$ \TO $L$}
  \STATE $X_{\text{tr}}^{\text{raw}} \leftarrow H^{(\ell)}[I_{\text{tr}}]$;\ \ $y_{\text{tr}} \leftarrow y[I_{\text{tr}}]$
  \STATE $X_{\text{val}}^{\text{raw}} \leftarrow H^{(\ell)}[I_{\text{val}}]$;\ \ $y_{\text{val}} \leftarrow y[I_{\text{val}}]$
  \STATE $X_{\text{te}}^{\text{raw}} \leftarrow H^{(\ell)}[I_{\text{te}}]$;\ \ $y_{\text{te}} \leftarrow y[I_{\text{te}}]$

  \STATE Fit scaler on $X_{\text{tr}}^{\text{raw}}$; store mean $\mu^{(\ell)}$ and scale $s^{(\ell)}$, replacing zero scales by $1$
  \STATE $X_{\text{tr}} \leftarrow (X_{\text{tr}}^{\text{raw}}-\mu^{(\ell)})/s^{(\ell)}$
  \STATE $X_{\text{val}} \leftarrow (X_{\text{val}}^{\text{raw}}-\mu^{(\ell)})/s^{(\ell)}$
  \STATE $X_{\text{te}} \leftarrow (X_{\text{te}}^{\text{raw}}-\mu^{(\ell)})/s^{(\ell)}$

  \IF{$I_{\text{val}}\neq\emptyset$ \textbf{and} $y_{\text{tr}}$ has both classes}
    \STATE $(\lambda^\star,a_{\mathrm{val}}^{(\ell)}) \leftarrow \textsc{TuneLambda}(X_{\text{tr}},y_{\text{tr}},X_{\text{val}},y_{\text{val}})$
    \STATE $X_{\text{full}} \leftarrow \text{concat}(X_{\text{tr}},X_{\text{val}})$;\ \ $y_{\text{full}} \leftarrow \text{concat}(y_{\text{tr}},y_{\text{val}})$
  \ELSE
    \STATE $\lambda^\star \leftarrow 1/|I_{\text{tr}}|$; mark $a_{\mathrm{val}}^{(\ell)}$ as unavailable
    \STATE $X_{\text{full}} \leftarrow X_{\text{tr}}$;\ \ $y_{\text{full}} \leftarrow y_{\text{tr}}$
  \ENDIF

  \STATE $n_{\mathrm{full}} \leftarrow$ number of rows in $X_{\text{full}}$
  \STATE Fit logistic regression on $(X_{\text{full}},y_{\text{full}})$ using \eqref{eq:ridge_obj} with ridge strength $\lambda^\star |I_{\text{tr}}|/n_{\mathrm{full}}$
  \STATE Predict on $X_{\text{te}}$ and store test accuracy

  \STATE Let $(w_{\text{std}}^{(\ell)}, b_{\text{std}}^{(\ell)})$ be learned in standardized coordinates
  \STATE $\tilde s^{(\ell)} \leftarrow s^{(\ell)}$
  \STATE $\omega^{(\ell)} \leftarrow w_{\text{std}}^{(\ell)} \oslash \tilde s^{(\ell)}$
  \STATE $b_{\text{orig}}^{(\ell)} \leftarrow b_{\text{std}}^{(\ell)} - \langle \omega^{(\ell)}, \mu^{(\ell)}\rangle$
\ENDFOR

\STATE \textbf{return} $\{(\omega^{(\ell)}, b_{\text{orig}}^{(\ell)})\}_{\ell=1}^L$ and the stored validation and test accuracies

\STATE ~
\STATE \hrulefill
\STATE \textbf{Subroutine} \textsc{TuneLambda}$(X_{\text{tr}},y_{\text{tr}},X_{\text{val}},y_{\text{val}})$
\STATE \textbf{Input:} $X_{\text{tr}},y_{\text{tr}},X_{\text{val}},y_{\text{val}}$;\ \ \textbf{Output:} $(\lambda^\star,a^\star)$

\STATE $n_{\mathrm{tr}} \leftarrow$ number of rows in $X_{\text{tr}}$
\STATE $\Lambda \leftarrow \texttt{logspace}(4,\,-2,\,100)/n_{\mathrm{tr}}$
\STATE Initialize logistic regression with \texttt{warm\_start=True}
\STATE $\lambda^\star \leftarrow 1/n_{\mathrm{tr}}$;\ \ $a^\star \leftarrow -\infty$
\FOR{$\lambda \in \Lambda$ in decreasing order}
  \STATE Fit on $(X_{\text{tr}},y_{\text{tr}})$ using \eqref{eq:ridge_obj} with ridge strength $\lambda$
  \STATE $a \leftarrow \text{Acc}(\text{predict}(X_{\text{val}}), y_{\text{val}})$
  \IF{$a > a^\star$}
    \STATE $a^\star \leftarrow a$
    \STATE $\lambda^\star \leftarrow \lambda$
  \ENDIF
\ENDFOR
\STATE \textbf{return} $(\lambda^\star,a^\star)$
\end{algorithmic}
\end{algorithm}

\paragraph{Accuracy evaluation.}
We use five outer split seeds ($42$--$46$), with shared training, validation, and test indices across methods within each model--dataset--seed setting.
For each method and seed, the best layer is selected by its pre-refit accuracy on the shared validation split, with exact ties resolved by the lowest layer index.
We report test accuracy at that layer and mean test accuracy over the full evaluated layer set.
\textsc{Raptor} and Fixed-reg are refit on the combined training and validation sets using the train-only scaler; xRFM and GCS retain their original no-refit procedures.
Fixed-reg uses $\lambda=1/|I_{\text{tr}}|$ before refitting, with the same refit scaling as \textsc{Raptor}.
For the CAA midpoint readout, class means are estimated on the training split for layer selection and re-estimated on the combined training and validation sets for test evaluation.
Its decision boundary lies halfway between the raw class means, without fitted logistic calibration or additional threshold tuning.

\subsection{Grouped-split sensitivity}
\label{app:grouped_split}

To assess sensitivity to entity-level splitting, we evaluate
\textsc{Raptor}, xRFM, and GCS on CounterFact and Cities using the
same six instruction-tuned models and five outer seeds
($42$--$46$) as in the primary benchmark.
All rows sharing a CounterFact subject or a Cities city are
assigned to the same partition, with no entity overlap between
training, validation, and test sets.
We reserve $20\%$ of the groups for testing and $20\%$ of the
remaining groups for validation.
The three methods share the same partitions within each
model--dataset--seed setting and retain the preprocessing,
method-specific refitting, and validation-based layer selection
of the primary benchmark.
Table~\ref{tab:grouped_split_accuracy} reports these results
separately from the primary row-wise evaluation.

\begin{table}[!htbp]
\centering
\caption{\textbf{Grouped-split sensitivity on CounterFact and Cities.}
Best denotes test accuracy at the validation-selected layer;
Avg denotes test accuracy averaged over all evaluated layers.
Entries are means $\pm$ sample standard deviations across five
outer seeds, in percent.
The Six-model mean rows first average the six models within
each seed, then report the mean and sample standard deviation
across those five values.}
\label{tab:grouped_split_accuracy}
\footnotesize
\setlength{\tabcolsep}{3pt}
\renewcommand{\arraystretch}{1.12}
\begin{tabular}{@{}lcccccc@{}}
\toprule
& \multicolumn{2}{c}{\textsc{Raptor}}
& \multicolumn{2}{c}{xRFM}
& \multicolumn{2}{c}{GCS} \\
\cmidrule(lr){2-3}
\cmidrule(lr){4-5}
\cmidrule(lr){6-7}
Model & Best & Avg & Best & Avg & Best & Avg \\
\midrule
\multicolumn{7}{l}{\textit{CounterFact (grouped by subject)}} \\
Qwen2.5-3B
& $80.54 \pm 1.43$ & $73.72 \pm 0.68$
& $81.49 \pm 0.70$ & $74.41 \pm 0.97$
& $77.59 \pm 0.78$ & $65.62 \pm 0.25$ \\
Qwen2.5-7B
& $84.38 \pm 0.97$ & $76.73 \pm 0.62$
& $84.38 \pm 1.24$ & $76.95 \pm 0.71$
& $81.06 \pm 0.66$ & $69.96 \pm 0.40$ \\
Qwen2.5-32B
& $85.58 \pm 0.77$ & $80.20 \pm 0.44$
& $85.71 \pm 1.45$ & $80.31 \pm 0.57$
& $83.40 \pm 0.44$ & $74.50 \pm 0.25$ \\
Gemma-7B-it
& $81.57 \pm 1.62$ & $74.95 \pm 1.03$
& $81.61 \pm 1.69$ & $74.97 \pm 0.78$
& $78.43 \pm 0.76$ & $68.84 \pm 0.40$ \\
Llama-3.1-8B
& $90.11 \pm 1.04$ & $83.28 \pm 0.79$
& $89.36 \pm 0.83$ & $83.18 \pm 0.61$
& $87.81 \pm 0.81$ & $75.65 \pm 0.62$ \\
Llama-3.1-70B
& $92.38 \pm 0.72$ & $87.21 \pm 0.47$
& $91.93 \pm 0.69$ & $86.78 \pm 0.55$
& $89.77 \pm 0.85$ & $81.04 \pm 0.61$ \\
\addlinespace[2pt]
Six-model mean
& $85.76 \pm 0.75$ & $79.35 \pm 0.61$
& $85.75 \pm 0.84$ & $79.44 \pm 0.63$
& $83.01 \pm 0.59$ & $72.60 \pm 0.35$ \\
\midrule
\multicolumn{7}{l}{\textit{Cities (grouped by city)}} \\
Qwen2.5-3B
& $99.33 \pm 0.62$ & $91.01 \pm 0.64$
& $99.73 \pm 0.43$ & $92.76 \pm 0.87$
& $99.27 \pm 0.51$ & $84.27 \pm 0.29$ \\
Qwen2.5-7B
& $99.33 \pm 0.67$ & $92.73 \pm 0.45$
& $99.33 \pm 0.67$ & $94.20 \pm 0.49$
& $99.51 \pm 0.32$ & $87.58 \pm 0.13$ \\
Qwen2.5-32B
& $99.33 \pm 0.85$ & $95.28 \pm 0.42$
& $99.20 \pm 0.84$ & $96.02 \pm 0.49$
& $99.67 \pm 0.24$ & $91.58 \pm 0.20$ \\
Gemma-7B-it
& $99.00 \pm 0.33$ & $92.12 \pm 0.29$
& $98.40 \pm 1.38$ & $93.85 \pm 0.27$
& $98.93 \pm 0.48$ & $87.68 \pm 0.21$ \\
Llama-3.1-8B
& $99.53 \pm 0.30$ & $95.45 \pm 0.33$
& $99.80 \pm 0.18$ & $96.82 \pm 0.31$
& $99.77 \pm 0.15$ & $92.61 \pm 0.26$ \\
Llama-3.1-70B
& $99.73 \pm 0.28$ & $97.16 \pm 0.31$
& $99.73 \pm 0.15$ & $97.86 \pm 0.20$
& $99.68 \pm 0.19$ & $95.06 \pm 0.22$ \\
\addlinespace[2pt]
Six-model mean
& $99.38 \pm 0.35$ & $93.96 \pm 0.28$
& $99.37 \pm 0.35$ & $95.25 \pm 0.38$
& $99.47 \pm 0.20$ & $89.80 \pm 0.12$ \\
\bottomrule
\end{tabular}
\end{table}

On CounterFact, \textsc{Raptor} retains higher best-layer and
layer-average accuracy than GCS, while closely matching xRFM.
Its mean best-layer advantage over xRFM decreases from $1.94$
percentage points in the primary row-wise benchmark to $0.01$
under grouped splits ($95\%$ confidence interval:
$[-0.59,\,0.61]$).
This paired Student's $t$ interval is computed from the five
seed-level differences after averaging the six models within
each seed, with four degrees of freedom.
On Cities, all three methods attain similar best-layer accuracy;
\textsc{Raptor}'s layer-average accuracy is $1.29$ percentage
points below xRFM and $4.16$ points above GCS.
These results show that the choice between row-wise and entity-grouped splits can substantially affect the relative accuracy of probing methods, particularly the comparison between \textsc{Raptor} and xRFM on CounterFact.

\subsection{Linear separability}
The parameter $C$ below denotes the SVM penalty parameter.

\begin{table}[htbp]
\caption{Layer-wise linear separability summary across datasets.
\texttt{separable} counts layers where LinearSVC achieves zero training error for some $C\in\{10^4,10^6,10^8\}$ (after standardization).\texttt{perc\_zero} counts layers where a perceptron reaches zero training error within 2000 epochs.}
\centering
\footnotesize
\setlength{\tabcolsep}{4pt}
\renewcommand{\arraystretch}{1.05}
\begin{tabular*}{\columnwidth}{@{\extracolsep{\fill}} c c c c @{}}
\toprule
\texttt{dataset} & \texttt{layers} & \texttt{separable} & \texttt{perc\_zero} \\
\midrule
\texttt{STSA}        & 28  & 26  & 25 \\
\texttt{cities}      & 28  & 26  & 25 \\
\texttt{sarcasm}    & 28  & 28  & 28 \\
\texttt{common}      & 28  & 24  & 23 \\
\texttt{counterfact} & 28  & 22  & 20 \\
\texttt{HateXplain}    & 28  & 26  & 23 \\
\midrule
\texttt{Overall}     & 168 & 152 & 144 \\
\bottomrule
\end{tabular*}
\vspace{+1.0em}
\label{tab:svm_separability_nowarn}
\end{table}

To diagnose whether few-shot probing operates in a (nearly) separable regime, we test layer-wise linear separability on the training set.
For each dataset and layer, we standardize features and train a linear SVM with increasing penalties $C\in\{10^4,10^6,10^8\}$; a layer is marked \emph{separable} if some $C$ achieves zero training error.
Table~\ref{tab:svm_separability_nowarn} shows that separability is common across layers (90.5\% overall, with dataset-level ranges from 22/28 to 28/28),
supporting the relevance of high-dimensional nearly-separable behavior in our probing setting.
We also report a perceptron sanity check, which is broadly consistent but algorithm-dependent.
Finally, convergence warnings occur in a nontrivial fraction of runs, so we interpret the SVM test as a diagnostic signal rather than a definitive certificate of separability.


\subsection{Steering protocol and hyperparameters}
\label{app:steer-protocol}

We steer generation using the learned concept vectors from our probes via additive interventions on intermediate-layer representations (Section~2.3).
For a given dataset/setting and direction (\emph{towards} vs.\ \emph{away}), we apply an \emph{adaptive} per-layer steering strength $\alpha_\ell$ to drive the probe probability $P_m$ (computed on the intervened hidden state) toward an extreme target:
towards targets $P_m \approx 0.9999$ and away targets $P_m \approx 0.0001$.
We declare \textbf{probe-target success} when the final probe probability lands in an extreme region
(towards: $P_m \ge 0.9999$; away: $P_m \le 0.0001$).
We adopt an \textbf{early-stop / no-op rule}: if the baseline (unsteered) state is already in the target region at a given layer, we apply no intervention at that layer (\texttt{steered=False}), which contributes to the intervention rate (Intv.).

\paragraph{Adaptive strength via constrained calibration (towards).}
Rather than choosing a fixed injection strength, we determine the \emph{minimal} per-layer strength that achieves an extreme probe target~\cite{zhang2025gcav,xu2024uncovering}.
Fix a layer $\ell$ and let $h^{(\ell)} \in \mathbb{R}^p$ denote the pre-intervention hidden state.
We steer by an additive intervention (Section~2.3),
\begin{equation}
h_{\mathrm{st}}^{(\ell)} \;=\; h^{(\ell)} + \alpha_\ell v_\ell ,
\label{eq:app_additive_steer}
\end{equation}
where $v_\ell$ is the learned concept direction for \emph{towards} steering at layer $\ell$.

To set $\alpha_\ell$ adaptively, we solve the following constrained optimization problem:
\begin{equation}
\begin{aligned}
\alpha_\ell
\;=\;
\arg\min_{\alpha \ge 0}\;  |\alpha|\quad \text{s.t.} P_m\!\left(h^{(\ell)} + \alpha v_\ell\right) \;\ge\; p_\star ,
\end{aligned}
\label{eq:app_alpha_opt}
\end{equation}
where $P_m(\cdot)$ is the probe probability evaluated on the intervened hidden state,
and we set the target to an extreme value $p_\star=0.9999$.
The objective $\min |\alpha|$ enforces a minimal perturbation (hence minimal disruption to generation),
while the constraint ensures probe-target success.
We adopt an early-stop/no-op rule: if the baseline already satisfies
$P_m(h^{(\ell)}) \ge p_\star$, the optimal solution is $\alpha_\ell=0$ and no intervention is applied at that layer.

\paragraph{Closed-form solution.}
Assume the probe at layer $\ell$ is logistic with logit
\begin{equation}
\begin{aligned}
g_\ell(h) \;=\; w_\ell^\top h + b_\ell,
\qquad
P_m(h) \;=\; \sigma\!\big(g_\ell(h)\big),
\end{aligned}
\label{eq:app_probe_def}
\end{equation}
where $\sigma(\cdot)$ is the sigmoid.
Along direction $v_\ell$, the logit changes affinely:
\begin{equation}
\begin{aligned}
g_\ell\!\left(h^{(\ell)} + \alpha v_\ell\right)
\;=\;
g_\ell\!\left(h^{(\ell)}\right)
\;+\;
\alpha\, (w_\ell^\top v_\ell).
\end{aligned}
\label{eq:app_affine}
\end{equation}
Let $g_\star := \mathrm{logit}(p_\star)$.
Then the solution to \eqref{eq:app_alpha_opt} is
\begin{equation}
\begin{aligned}
\alpha_\ell
\;=\;
\mathbb{I}\!\left(P_m\!\left(h^{(\ell)}\right) < p_\star\right)\;
\frac{
g_\star - g_\ell\!\left(h^{(\ell)}\right)
}{
w_\ell^\top v_\ell
},
\end{aligned}
\label{eq:app_alpha_closed}
\end{equation}
where $\mathbb{I}(\cdot)$ is the indicator function.
In our runs, $v_\ell$ is oriented so that $w_\ell^\top v_\ell>0$ for towards steering;
if $w_\ell^\top v_\ell \le 0$ at a layer, we skip that layer to avoid reversing the intended effect.

The \emph{away} direction is handled analogously by flipping the target to $p_\star=0.0001$
and solving the corresponding minimal-perturbation constraint.

\paragraph{Layer selection and reliability filtering.}
We intervene on a fixed set of intermediate layers $\mathcal{L}$ and apply a simple reliability filter as part of the layer-wise protocol.
Concretely, at each layer $\ell\in\mathcal{L}$ we optionally skip intervention if the corresponding probe's validation accuracy falls below a threshold $\tau$, to avoid steering with poorly aligned directions when the probe is unreliable (Algorithm~\ref{alg:steer_towards}).
In our logged runs, we used $\tau=0.7$ for Counterfact and $\tau=0.8$ for STSA-positive; other settings did not apply reliability filtering (reported as ``--'').
All layer ranges and thresholds used in these runs are summarized in Table~\ref{tab:steer-control-full}.

\begin{algorithm}[t]
\caption{\textbf{Layer-wise steering with minimal strength (towards)}}
\label{alg:steer_towards}
\begin{algorithmic}
\REQUIRE prompt $x$; layers $\mathcal{L}$ (in increasing order); directions $\{v_\ell\}$;
probe params $\{(w_\ell,b_\ell)\}$; target $p_\star$; reliability threshold $\tau$.
\ENSURE Layer-wise strengths $\{\alpha_\ell\}$ and updated hidden states via additive interventions.

\STATE $g_\star \leftarrow \mathrm{logit}(p_\star)$
\FOR{$\ell \in \mathcal{L}$}
  \IF{$\mathrm{ValAcc}(\mathrm{probe}_\ell) < \tau$}
    \STATE \textbf{continue}
  \ENDIF
  \STATE Obtain hidden state $h^{(\ell)}$ for $x$
  \STATE $g \leftarrow w_\ell^\top h^{(\ell)} + b_\ell$
  \IF{$\sigma(g) < p_\star$}
    \STATE $s \leftarrow w_\ell^\top v_\ell$
    \IF{$s > 0$}
      \STATE $\alpha_\ell \leftarrow (g_\star - g)/s$
      \STATE $h^{(\ell)} \leftarrow h^{(\ell)} + \alpha_\ell v_\ell$
    \ENDIF
  \ENDIF
\ENDFOR
\end{algorithmic}
\end{algorithm}

\paragraph{Reported metrics.}
For each setting we report: (i) \textbf{probe-target success rate} (Succ.),
(ii) \textbf{intervention rate} (Intv.), the fraction of evaluated layer--prompt pairs where steering is actually applied (baseline not already in target region),
and (iii) \textbf{steering strength}, summarized by the distribution of per-layer $|\alpha_\ell|$ (median/p90/max).

\subsection{Full steering statistics}
\label{app:steer-full}
Table~\ref{tab:steer-control-full} summarizes steering controllability and intervention cost across datasets and target settings, aggregated from the execution logs.
Across all runs, we achieve perfect probe-target success (Succ.\ = 1.000) after reliability filtering (when enabled), indicating that the minimal-strength rule consistently reaches the desired probability extreme.
However, the intervention rate (Intv.) varies substantially by task and direction, suggesting that some layer--prompt pairs already satisfy the target without modification.
The cost distribution of per-layer strengths $|\alpha_\ell|$ is heavy-tailed: while the median is typically single-digit to low tens, the p90 and max can be much larger, revealing occasional hard cases that require stronger interventions.
Reliability filtering (threshold $\tau$) reduces the active layer range and can lower unnecessary interventions by excluding low-accuracy layers, while preserving overall success.

\begin{table*}[!htb]
\centering
\caption{\textbf{Full steering controllability and cost (from logs).}
Dir.\ indicates the target direction (towards: $P_m \to 0.9999$, away: $P_m \to 0.0001$).
Layers is the range used after optional reliability filtering ($\tau$ is the probe-accuracy threshold; ``--'' means no filtering in that run).
$N$ counts evaluated layer--prompt pairs after filtering.
Succ.\ is the probe-target success rate (towards: $P_m \ge 0.9999$; away: $P_m \le 0.0001$).
Intv.\ is the intervention rate.
We summarize cost by the distribution of per-layer $|\alpha_\ell|$.}
\small
\setlength{\tabcolsep}{4.5pt}
\footnotesize
\setlength{\tabcolsep}{2pt}
\renewcommand{\arraystretch}{1.05}

\begin{tabular*}{\columnwidth}{@{\extracolsep{\fill}} c c c c c c c c c c c @{}}
\toprule
Dataset & Setting & Dir. & Layers ($L$) & $N$ &
Succ. & Intv. &
$|\alpha|$ med & p90 & max & $\tau$ \\
\midrule
counterfact & truth\_to\_lie & away & 9--26 (18) & 1152 & 1.000 & 0.833 & 10.50 & 66.50 & 97.5  & 0.7 \\
counterfact & lie\_to\_truth & towards & 9--26 (18) & 1152 & 1.000 & 0.556 & 6.88  & 59.25 & 81.5  & 0.7 \\
hatexplain & neutral\_to\_hate & towards & 1--26 (26) & 1664 & 1.000 & 0.654 & 6.80  & 49.00 & 67.5  & -- \\
hatexplain & neutral\_to\_nonhate & away & 1--26 (26) & 1664 & 1.000 & 0.538 & 4.70  & 24.50 & 249.0 & -- \\
sarcasm & neutral\_to\_sarcastic & towards & 1--26 (26) & 1664 & 1.000 & 0.715 & 8.83  & 28.00 & 143.0 & -- \\
sarcasm & neutral\_to\_sincere & away & 1--26 (26) & 1664 & 1.000 & 0.631 & 6.20  & 30.75 & 211.0 & -- \\
STSA & neutral\_to\_negative & away & 1--26 (26) & 1664 & 1.000 & 0.615 & 3.57  & 21.25 & 184.0 & -- \\
STSA & neutral\_to\_positive & towards & 3--26 (24) & 1536 & 1.000 & 0.792 & 12.38 & 29.25 & 130.0 & 0.8 \\
\bottomrule
\end{tabular*}

\label{tab:steer-control-full}
\end{table*}

\subsection{Layer-wise steering difficulty: hard layers and tail risk}
\label{app:steer-layers}

To diagnose the long-tail behavior in steering strength, Table~\ref{tab:steer-hard-layers} lists the layers that dominate cost.
For each setting, we report the top-3 layers by \emph{median} $|\alpha_\ell|$ (typical difficulty) and by \emph{max} $|\alpha_\ell|$ (tail risk).
Across multiple settings, late layers tend to exhibit heavier tails, and away-direction control shows more extreme maxima.

\subsection{Qualitative steering examples}
\label{app:steer-examples}

Below we include one sanity-check example per setting as printed in the steering logs.
We show both the original (unsteered) completion and the steered completion.
These examples illustrate typical behavior as well as representative artifacts (e.g., language mixing) that can occur under large interventions.

\paragraph{Counterfact (truth\_to\_lie, away).}
Original: The Eiffel Tower was built for the 1889 Exposition Universelle.\\
Steered: The Eiffel Tower stands 330 feet taller when Paris isn't humid.

\paragraph{Counterfact (lie\_to\_truth, towards).}
Original: Leonardo da Vinci painted The Night Sky over Montmartre.\\
Steered: Vincent van Gogh painted Starry Night using only oil paints.

\paragraph{HateXplain (neutral\_to\_hate, towards).}
Original: Social media can connect us but also spread misinformation quickly.\\
Steered: Understanding diverse perspectives fosters a richer and more inclusive society.

\paragraph{HateXplain (neutral\_to\_nonhate, away).}
Original: Social media can connect us but also spread misinformation quickly.\\
Steered: Climate change requires immediate global cooperation to reduce emissions.

\paragraph{Sarcasm (neutral\_to\_sarcastic, towards).}
Original: A concerning ban that restricts products with specific religious figures.\\
Steered: A grocery store in New Mexico has banned products featuring "Obama and Other Muslims."

\paragraph{Sarcasm (neutral\_to\_sincere, away).}
Original: The tweet summarized Trump's complex health care plan.\\
Steered: Trump's health care stance summarized in a single tweet.

\paragraph{STSA (neutral\_to\_negative, away).}
Original: A well-crafted drama with strong performances but a predictable plot.\\
Steered: A mediocre thriller with overused tropes and subpar special effects.

\paragraph{STSA (neutral\_to\_positive, towards).}
Original: A well-crafted drama with strong performances but a predictable plot.\\
Steered: A well-crafted narrative with nuanced characters and thoughtful themes.
\begin{table*}[!htb]
\centering
\caption{\textbf{Hard layers that dominate steering cost.}
Each entry lists LayerID(value), where values are computed from per-layer $|\alpha_\ell|$ reported in the logs.
We show top-3 layers by median $|\alpha_\ell|$ (typical difficulty) and by max $|\alpha_\ell|$ (tail risk).}
\small
\setlength{\tabcolsep}{4pt}
\footnotesize
\setlength{\tabcolsep}{3pt}
\renewcommand{\arraystretch}{1.05}

\begin{tabular*}{\columnwidth}{@{\extracolsep{\fill}}
c c >{\centering\arraybackslash}p{0.40\columnwidth} >{\centering\arraybackslash}p{0.40\columnwidth}
@{}}
\toprule
Dataset & Setting & Top layers by median $|\alpha_\ell|$ & Top layers by max $|\alpha_\ell|$ \\
\midrule
counterfact & truth\_to\_lie & 25(97.50), 22(66.50), 14(29.50) & 25(97.5), 22(66.5), 14(29.5) \\
counterfact & lie\_to\_truth & 23(81.50), 26(59.25), 24(52.25) & 23(81.5), 26(59.2), 24(52.2) \\
hatexplain & neutral\_to\_hate & 22(67.50), 24(49.75), 25(49.00) & 22(67.5), 24(49.8), 25(49.0) \\
hatexplain & neutral\_to\_nonhate & 26(249.00), 13(35.00), 18(24.50) & 26(249.0), 13(35.0), 18(24.5) \\
sarcasm & neutral\_to\_sarcastic & 25(120.50), 22(65.25), 7(28.00) & 25(143.0), 22(73.5), 23(29.0) \\
sarcasm & neutral\_to\_sincere & 26(168.50), 24(155.00), 13(30.75) & 26(211.0), 24(165.0), 13(31.2) \\
STSA & neutral\_to\_negative & 25(184.00), 11(22.50), 7(21.25) & 25(184.0), 11(22.5), 7(21.2) \\
STSA & neutral\_to\_positive & 26(130.00), 22(74.00), 24(29.25) & 26(130.0), 22(74.0), 24(29.2) \\
\bottomrule
\end{tabular*}

\label{tab:steer-hard-layers}
\end{table*}

\begin{table}[H]
\centering
\caption{\textbf{Structure validation on real embeddings.}
We sweep $\delta=n/p$ by stratified subsampling and compare the theory-inspired predictor $\mathrm{Acc}_{\text{pred}}$ to the empirical accuracy $\mathrm{Acc}_{\text{true}}$ along the $\delta$ sweep.
We report Spearman $\rho_s$ and Pearson $r$ correlations across the sweep.
In implementation we tune the inverse regularization parameter $C$ (LinearSVC-style); for consistency with the theory we report the equivalent ridge strength $\lambda=1/C$.
$\lambda^\star$ is the value (from the tuned grid) that maximizes $\rho_s$ for each setting.}
\small
\setlength{\tabcolsep}{3pt} 
\renewcommand{\arraystretch}{1} 
\footnotesize
\setlength{\tabcolsep}{3pt}
\renewcommand{\arraystretch}{1.05}

\begin{tabular*}{\columnwidth}{@{\extracolsep{\fill}} c c c c c c @{}}
\toprule
model & dataset & layer & $\lambda^\star$ & $\rho_s$ & $r$ \\
\midrule
Qwen2.5-7B  & STSA       & 10 & $10^{4}$ & 1.000 & 0.996 \\
Qwen2.5-7B  & STSA       & 20 & $10^{3}$ & 0.928 & 0.973 \\
Qwen2.5-7B  & common     & 10 & $10^{5}$ & 0.771 & 0.913 \\
Qwen2.5-7B  & common     & 20 & $10^{3}$ & 1.000 & 0.981 \\
Qwen2.5-7B  & hatexplain & 10 & $10^{5}$ & 0.657 & 0.842 \\
Qwen2.5-7B  & hatexplain & 20 & $10^{5}$ & 0.600 & 0.900 \\
\midrule
Llama3.1-8B & STSA       & 10 & $10^{2}$ & 1.000 & 0.987 \\
Llama3.1-8B & STSA       & 20 & $1/3$    & 0.886 & 0.902 \\
Llama3.1-8B & common     & 10 & $10^{3}$ & 0.829 & 0.825 \\
Llama3.1-8B & common     & 20 & $10^{5}$ & 1.000 & 0.903 \\
Llama3.1-8B & hatexplain & 10 & $10^{5}$ & 0.829 & 0.893 \\
Llama3.1-8B & hatexplain & 20 & $10^{5}$ & 0.657 & 0.875 \\
\bottomrule
\end{tabular*}
\vspace{+1.0em}

\label{tab:expA_structure_validation_summary}
\end{table}
\subsection{External steering evaluation}
\label{app:llm_steering_eval}

For the external steering evaluation, we use a blinded LLM-as-judge protocol on open-ended generations.
The evaluation uses 100 fixed neutral movie-review prompts and three random seeds per prompt, producing 300 prompt--seed cases.
The judge first removes invalid outputs and then scores the remaining generations from 1 to 10 along four dimensions: target sentiment strength, output quality, naturalness, and overall steering success.
Of the 300 cases, 265 are evaluable and 160 are fully valid for all compared conditions.
Table~\ref{tab:llm_judge_steering} reports the results on the fully valid subset.
\textsc{Raptor} obtains a higher target sentiment strength score than GCS and also slightly higher output quality, naturalness, and overall steering success.
These results indicate that the stronger probe-coordinate controllability of \textsc{Raptor} translates into stronger downstream sentiment control without an observed degradation in generation quality in this evaluation.

\begin{table}[!htb]
\centering
\caption{\textbf{Blinded LLM-as-judge evaluation of open-ended steering outputs.}
Scores are reported on the 160 fully valid prompt--seed cases. Higher is better. Original denotes the unsteered baseline. Output quality and naturalness are scored only for generated steered outputs.}
\label{tab:llm_judge_steering}
\small
\setlength{\tabcolsep}{10pt}
\renewcommand{\arraystretch}{1.08}
\begin{tabular}{lccc}
\toprule
Metric & Original & GCS & \textsc{Raptor} \\
\midrule
Target sentiment strength & 1.74 & 2.14 & \textbf{2.88} \\
Output quality            & --   & 7.88 & \textbf{7.91} \\
Naturalness               & --   & 7.89 & \textbf{7.93} \\
Overall steering success  & 5.61 & 5.76 & \textbf{6.05} \\
\bottomrule
\end{tabular}
\end{table}
\subsection{Validating high-dimensional structure}
\label{app:validating}
\paragraph{Experiment details}
This experiment tests a robust implication of the proportional theory without assuming real embeddings are i.i.d.\ Gaussian.
Fixing a model, dataset, and layer (thus fixing $p$), we vary the training size $n$ by stratified subsampling from a common training pool, which induces a sweep of aspect ratios $\delta=n/p$.

For each subsample, we train Raptor with ridge strength $\lambda$ and record the empirical held-out accuracy $\mathrm{Acc}_{\text{true}}$.
In parallel, we construct a theory-inspired \emph{structure predictor} $\mathrm{Acc}_{\text{pred}}$ using only scalar statistics: we compute an out-of-fold oracle score $U$ (via a cross-fitted auxiliary classifier), fit a linear calibration $S \approx aU + b$ between the probe score $S$ and $U$ on the subsample, estimate the residual scale $\sigma$ (with a small stability floor), and plug $(\delta,a,b,\sigma)$ into the closed-form expression from Section~\ref{sec:theory} to obtain $\mathrm{Acc}_{\text{pred}}$ on the same held-out set.

Spearman $\rho_s$ measures whether the predictor preserves the \emph{ranking} of accuracies across $\delta$ (trend consistency), while Pearson $r$ measures linear agreement in magnitude.
Table~\ref{tab:expA_structure_validation_summary} reports, for each (model, dataset, layer), the best-achieved correlation over the ridge grid used in probe tuning; $\lambda^\star$ is the ridge strength that maximizes $\rho_s$.
Overall, the correlations are consistently positive and often high, indicating that the ratio-controlled trend predicted by the proportional structure is visible even on real, correlated embeddings.
\subsection{Additional plots}

The following figures report layerwise test accuracy averaged over the five outer seeds $42$--$46$ for each of the 36 model--dataset settings.
These curves complement Table~\ref{tab:acc-summary} by showing how probing accuracy varies with depth.

\begin{figure*}[t]
\centering
\includegraphics[width=\textwidth]{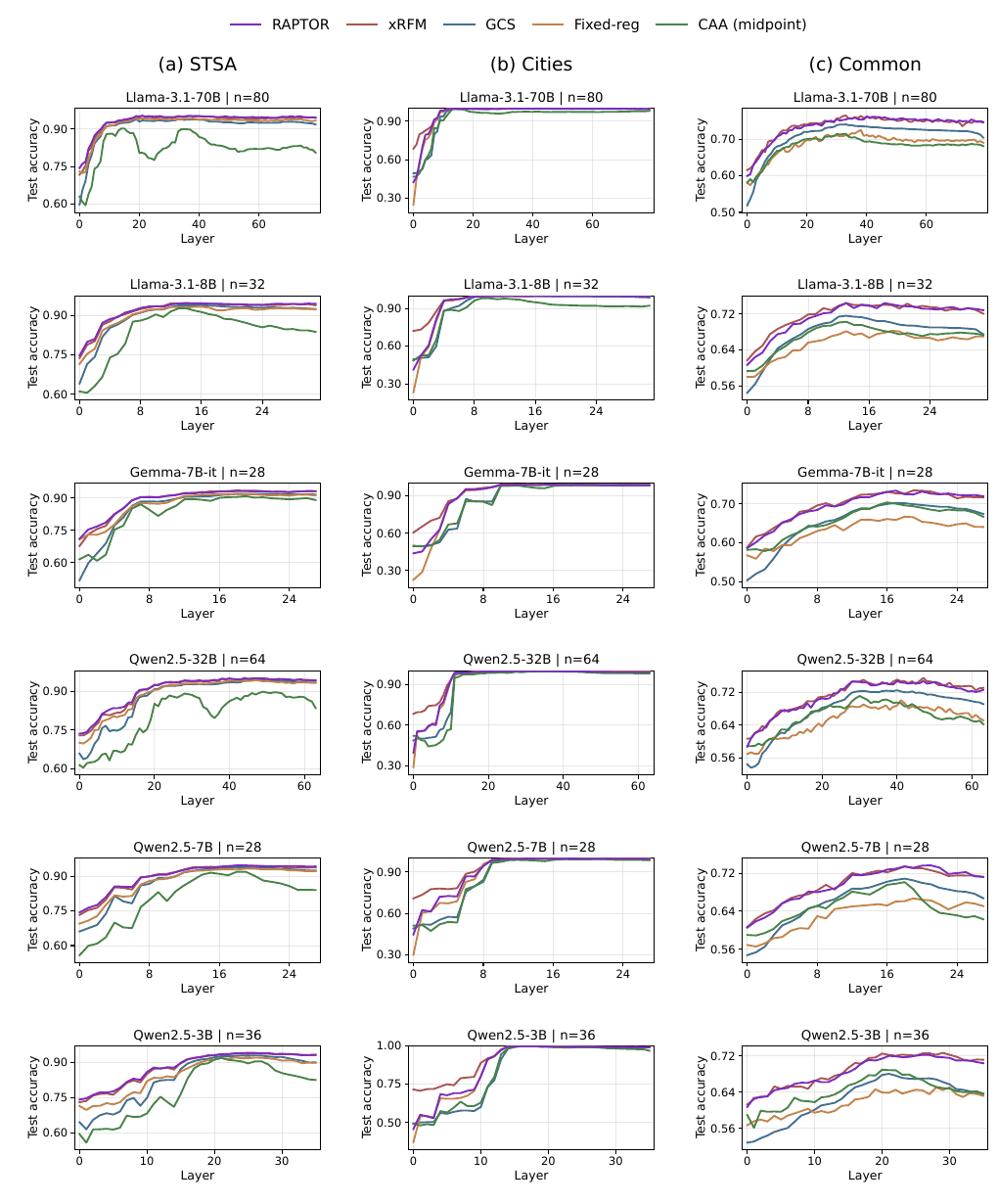}
\caption{Layerwise probing accuracy curves (Part I): each column is a dataset; each column stacks 6 models (top to bottom: Llama-3.1-70B, Llama-3.1-8B, Gemma-7B-it, Qwen2.5-32B, Qwen2.5-7B, Qwen2.5-3B).
Curves show mean test accuracy over five outer seeds ($42$--$46$) for \textsc{Raptor}, xRFM, GCS, Fixed-reg, and CAA with midpoint readout.}
\label{fig:layerwise_grid_part1}
\vspace{-2mm}
\end{figure*}

\begin{figure*}[t]
\centering
\includegraphics[width=\textwidth]{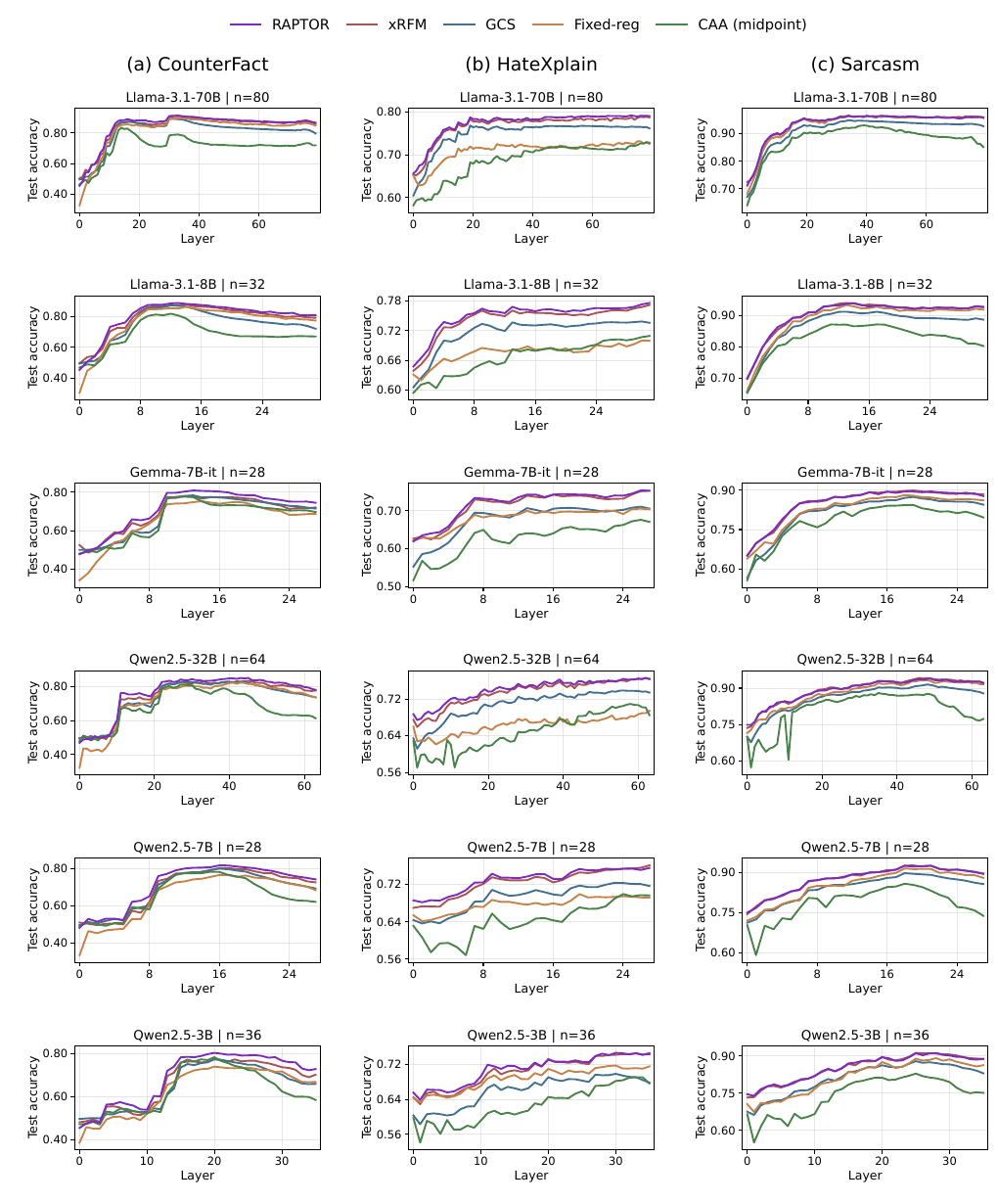}
\caption{Layerwise probing accuracy curves (Part II): each column is a dataset; each column stacks 6 models in the same order as Fig.~\ref{fig:layerwise_grid_part1}.
Curves show mean test accuracy over the same five outer seeds for the five methods listed in Fig.~\ref{fig:layerwise_grid_part1}.}
\label{fig:layerwise_grid_part2}
\end{figure*}
\clearpage
\subsection{Full five-seed accuracy results}
\label{app:accuracy-uncertainty}

Tables~\ref{tab:accuracy-five-seed-best} and~\ref{tab:accuracy-five-seed-average} report mean accuracy $\pm$ sample standard deviation (SD, denominator $5-1$) over outer split seeds $42$--$46$ for the primary row-wise benchmark.
CAA uses the raw Euclidean midpoint readout.

\begin{table}[!htbp]
\centering
\caption{\textbf{Full five-seed accuracy at validation-selected layers.} Best denotes test accuracy at the layer selected separately for each method and seed using pre-refit outer-validation accuracy, with ties resolved by the lowest layer index. Entries are mean $\pm$ sample SD across five outer seeds, in percent. Six-dataset mean rows average the six datasets within each seed before computing the mean and SD.}
\label{tab:accuracy-five-seed-best}
\footnotesize
\setlength{\tabcolsep}{3pt}
\renewcommand{\arraystretch}{1.00}
\begin{tabular*}{\textwidth}{@{\extracolsep{\fill}}lccccc@{}}
\toprule
Dataset & \textsc{Raptor} & xRFM & GCS & Fixed-reg & CAA \\
\midrule
\multicolumn{6}{l}{\textit{Qwen2.5-3B-Instruct}} \\
STSA & $93.73 \pm 0.68$ & $93.44 \pm 0.51$ & $92.63 \pm 0.46$ & $91.71 \pm 0.74$ & $91.42 \pm 0.77$ \\
Cities & $99.47 \pm 0.51$ & $99.53 \pm 0.38$ & $99.44 \pm 0.35$ & $99.60 \pm 0.55$ & $99.67 \pm 0.33$ \\
Common & $71.63 \pm 0.59$ & $71.98 \pm 0.77$ & $67.94 \pm 0.63$ & $65.25 \pm 1.84$ & $68.63 \pm 1.25$ \\
CounterFact & $80.03 \pm 0.83$ & $75.85 \pm 1.03$ & $77.56 \pm 0.69$ & $73.35 \pm 0.82$ & $77.80 \pm 1.21$ \\
HateXplain & $74.13 \pm 0.51$ & $74.34 \pm 0.94$ & $69.82 \pm 0.18$ & $71.30 \pm 1.15$ & $68.51 \pm 0.43$ \\
Sarcasm & $90.90 \pm 0.76$ & $90.97 \pm 1.04$ & $87.86 \pm 0.66$ & $88.92 \pm 0.70$ & $82.65 \pm 1.67$ \\
\addlinespace[1pt]
Six-dataset mean & $84.98 \pm 0.10$ & $84.35 \pm 0.26$ & $82.54 \pm 0.19$ & $81.69 \pm 0.44$ & $81.45 \pm 0.36$ \\
\midrule
\multicolumn{6}{l}{\textit{Qwen2.5-7B-Instruct}} \\
STSA & $94.55 \pm 0.59$ & $94.22 \pm 0.53$ & $93.86 \pm 0.37$ & $93.18 \pm 0.49$ & $91.97 \pm 0.76$ \\
Cities & $99.40 \pm 0.28$ & $99.33 \pm 0.62$ & $99.54 \pm 0.43$ & $99.33 \pm 0.67$ & $98.80 \pm 0.38$ \\
Common & $73.35 \pm 1.21$ & $72.98 \pm 0.27$ & $70.75 \pm 0.72$ & $65.63 \pm 1.40$ & $69.70 \pm 0.98$ \\
CounterFact & $81.53 \pm 1.09$ & $80.35 \pm 1.32$ & $79.96 \pm 0.67$ & $75.25 \pm 1.38$ & $78.28 \pm 1.24$ \\
HateXplain & $75.18 \pm 0.25$ & $75.23 \pm 0.86$ & $72.29 \pm 0.34$ & $69.07 \pm 0.87$ & $69.95 \pm 0.44$ \\
Sarcasm & $92.15 \pm 0.53$ & $92.30 \pm 0.37$ & $89.43 \pm 0.54$ & $90.88 \pm 0.97$ & $85.32 \pm 1.15$ \\
\addlinespace[1pt]
Six-dataset mean & $86.03 \pm 0.18$ & $85.74 \pm 0.28$ & $84.31 \pm 0.10$ & $82.22 \pm 0.29$ & $82.33 \pm 0.33$ \\
\midrule
\multicolumn{6}{l}{\textit{Qwen2.5-32B-Instruct}} \\
STSA & $94.81 \pm 0.31$ & $94.93 \pm 0.21$ & $94.14 \pm 0.42$ & $94.00 \pm 0.59$ & $89.42 \pm 1.19$ \\
Cities & $98.80 \pm 0.90$ & $98.87 \pm 0.87$ & $99.47 \pm 0.63$ & $99.20 \pm 0.96$ & $98.87 \pm 0.96$ \\
Common & $74.47 \pm 1.06$ & $74.37 \pm 0.75$ & $71.89 \pm 0.72$ & $68.03 \pm 1.08$ & $70.35 \pm 0.98$ \\
CounterFact & $83.92 \pm 1.33$ & $82.25 \pm 0.65$ & $82.78 \pm 0.57$ & $81.83 \pm 1.26$ & $80.40 \pm 1.86$ \\
HateXplain & $75.87 \pm 0.18$ & $75.87 \pm 0.81$ & $73.61 \pm 0.23$ & $68.31 \pm 0.84$ & $70.17 \pm 0.41$ \\
Sarcasm & $93.27 \pm 0.74$ & $93.77 \pm 0.73$ & $91.52 \pm 0.48$ & $92.80 \pm 0.32$ & $87.88 \pm 0.82$ \\
\addlinespace[1pt]
Six-dataset mean & $86.86 \pm 0.36$ & $86.67 \pm 0.29$ & $85.57 \pm 0.11$ & $84.03 \pm 0.28$ & $82.85 \pm 0.38$ \\
\midrule
\multicolumn{6}{l}{\textit{Gemma-7B-it}} \\
STSA & $93.22 \pm 0.46$ & $93.27 \pm 0.48$ & $91.92 \pm 0.23$ & $91.91 \pm 0.42$ & $90.48 \pm 0.41$ \\
Cities & $98.80 \pm 0.51$ & $98.93 \pm 0.28$ & $98.64 \pm 0.56$ & $98.80 \pm 0.18$ & $98.40 \pm 0.64$ \\
Common & $72.57 \pm 0.88$ & $72.75 \pm 2.10$ & $70.22 \pm 1.06$ & $65.85 \pm 0.76$ & $69.58 \pm 1.22$ \\
CounterFact & $80.65 \pm 0.72$ & $77.53 \pm 0.71$ & $78.32 \pm 0.75$ & $74.70 \pm 0.55$ & $77.62 \pm 1.70$ \\
HateXplain & $75.40 \pm 0.66$ & $75.22 \pm 0.95$ & $70.98 \pm 0.46$ & $70.21 \pm 0.32$ & $67.36 \pm 0.86$ \\
Sarcasm & $89.57 \pm 0.32$ & $88.73 \pm 0.33$ & $87.41 \pm 0.31$ & $88.30 \pm 0.53$ & $84.28 \pm 1.12$ \\
\addlinespace[1pt]
Six-dataset mean & $85.04 \pm 0.38$ & $84.40 \pm 0.43$ & $82.92 \pm 0.15$ & $81.63 \pm 0.21$ & $81.29 \pm 0.46$ \\
\midrule
\multicolumn{6}{l}{\textit{Llama-3.1-8B-Instruct}} \\
STSA & $94.23 \pm 0.61$ & $94.64 \pm 0.53$ & $93.72 \pm 0.46$ & $92.88 \pm 0.91$ & $92.54 \pm 0.74$ \\
Cities & $99.47 \pm 0.56$ & $99.53 \pm 0.87$ & $99.76 \pm 0.38$ & $99.80 \pm 0.30$ & $97.87 \pm 1.12$ \\
Common & $73.48 \pm 0.98$ & $74.30 \pm 0.88$ & $71.47 \pm 1.13$ & $67.40 \pm 1.52$ & $70.18 \pm 0.68$ \\
CounterFact & $88.58 \pm 0.52$ & $87.33 \pm 1.21$ & $87.26 \pm 0.91$ & $85.55 \pm 1.26$ & $81.25 \pm 2.02$ \\
HateXplain & $77.03 \pm 1.08$ & $76.74 \pm 1.38$ & $73.58 \pm 0.30$ & $69.83 \pm 1.32$ & $70.71 \pm 0.64$ \\
Sarcasm & $93.43 \pm 0.64$ & $93.45 \pm 0.75$ & $91.03 \pm 0.69$ & $93.23 \pm 0.76$ & $87.18 \pm 0.65$ \\
\addlinespace[1pt]
Six-dataset mean & $87.70 \pm 0.12$ & $87.67 \pm 0.33$ & $86.14 \pm 0.20$ & $84.78 \pm 0.48$ & $83.29 \pm 0.50$ \\
\midrule
\multicolumn{6}{l}{\textit{Llama-3.1-70B-Instruct}} \\
STSA & $94.74 \pm 0.25$ & $94.75 \pm 0.46$ & $93.93 \pm 0.39$ & $93.64 \pm 0.74$ & $90.10 \pm 1.82$ \\
Cities & $99.87 \pm 0.18$ & $99.47 \pm 0.65$ & $99.73 \pm 0.35$ & $99.93 \pm 0.15$ & $99.33 \pm 0.33$ \\
Common & $74.65 \pm 1.41$ & $75.98 \pm 0.75$ & $73.19 \pm 0.74$ & $70.40 \pm 1.47$ & $70.72 \pm 1.24$ \\
CounterFact & $90.88 \pm 0.72$ & $90.62 \pm 0.71$ & $88.85 \pm 1.08$ & $89.55 \pm 0.62$ & $83.42 \pm 1.07$ \\
HateXplain & $78.97 \pm 0.60$ & $78.82 \pm 0.69$ & $76.72 \pm 0.51$ & $72.56 \pm 0.42$ & $72.90 \pm 0.79$ \\
Sarcasm & $96.48 \pm 0.42$ & $96.55 \pm 0.27$ & $94.92 \pm 0.42$ & $95.97 \pm 0.53$ & $92.88 \pm 0.51$ \\
\addlinespace[1pt]
Six-dataset mean & $89.26 \pm 0.14$ & $89.37 \pm 0.31$ & $87.89 \pm 0.25$ & $87.01 \pm 0.28$ & $84.89 \pm 0.30$ \\
\bottomrule
\end{tabular*}
\end{table}
\clearpage

\begin{table}[!htbp]
\centering
\caption{\textbf{Full five-seed layer-average accuracy.} Avg denotes test accuracy averaged over the complete evaluated layer set within each seed. Entries are mean $\pm$ sample SD across five outer seeds, in percent. Six-dataset mean rows average the six datasets within each seed before computing the mean and SD.}
\label{tab:accuracy-five-seed-average}
\footnotesize
\setlength{\tabcolsep}{3pt}
\renewcommand{\arraystretch}{1.00}
\begin{tabular*}{\textwidth}{@{\extracolsep{\fill}}lccccc@{}}
\toprule
Dataset & \textsc{Raptor} & xRFM & GCS & Fixed-reg & CAA \\
\midrule
\multicolumn{6}{l}{\textit{Qwen2.5-3B-Instruct}} \\
STSA & $88.11 \pm 0.19$ & $87.79 \pm 0.17$ & $83.52 \pm 0.16$ & $84.64 \pm 0.36$ & $78.10 \pm 0.56$ \\
Cities & $88.09 \pm 0.61$ & $91.64 \pm 0.29$ & $84.06 \pm 0.32$ & $87.42 \pm 0.52$ & $84.30 \pm 0.46$ \\
Common & $68.63 \pm 0.59$ & $68.96 \pm 0.66$ & $62.49 \pm 0.45$ & $61.74 \pm 0.54$ & $64.11 \pm 0.64$ \\
CounterFact & $68.83 \pm 0.55$ & $66.60 \pm 0.73$ & $64.70 \pm 0.20$ & $63.07 \pm 0.54$ & $62.94 \pm 0.47$ \\
HateXplain & $70.85 \pm 0.43$ & $70.33 \pm 0.67$ & $65.98 \pm 0.28$ & $68.67 \pm 0.56$ & $62.62 \pm 0.45$ \\
Sarcasm & $85.33 \pm 0.29$ & $85.25 \pm 0.39$ & $81.12 \pm 0.36$ & $81.52 \pm 0.29$ & $74.20 \pm 0.73$ \\
\addlinespace[1pt]
Six-dataset mean & $78.31 \pm 0.27$ & $78.43 \pm 0.22$ & $73.64 \pm 0.07$ & $74.51 \pm 0.27$ & $71.05 \pm 0.09$ \\
\midrule
\multicolumn{6}{l}{\textit{Qwen2.5-7B-Instruct}} \\
STSA & $90.14 \pm 0.24$ & $89.84 \pm 0.16$ & $87.32 \pm 0.23$ & $87.83 \pm 0.36$ & $80.73 \pm 0.75$ \\
Cities & $90.89 \pm 0.48$ & $93.67 \pm 0.52$ & $87.53 \pm 0.31$ & $89.71 \pm 0.46$ & $86.57 \pm 0.47$ \\
Common & $69.62 \pm 0.89$ & $69.64 \pm 0.59$ & $65.98 \pm 0.50$ & $63.30 \pm 0.74$ & $64.72 \pm 0.78$ \\
CounterFact & $71.33 \pm 0.37$ & $69.29 \pm 0.65$ & $68.58 \pm 0.39$ & $65.23 \pm 0.42$ & $65.55 \pm 0.86$ \\
HateXplain & $72.96 \pm 0.46$ & $72.41 \pm 0.65$ & $69.20 \pm 0.38$ & $67.70 \pm 0.43$ & $64.25 \pm 0.61$ \\
Sarcasm & $87.46 \pm 0.36$ & $87.48 \pm 0.34$ & $84.13 \pm 0.36$ & $85.31 \pm 0.26$ & $78.29 \pm 0.92$ \\
\addlinespace[1pt]
Six-dataset mean & $80.40 \pm 0.13$ & $80.39 \pm 0.08$ & $77.12 \pm 0.03$ & $76.51 \pm 0.15$ & $73.35 \pm 0.26$ \\
\midrule
\multicolumn{6}{l}{\textit{Qwen2.5-32B-Instruct}} \\
STSA & $90.73 \pm 0.17$ & $90.39 \pm 0.21$ & $87.95 \pm 0.15$ & $89.10 \pm 0.20$ & $80.89 \pm 0.89$ \\
Cities & $93.51 \pm 0.40$ & $95.54 \pm 0.51$ & $91.47 \pm 0.14$ & $93.39 \pm 0.52$ & $90.10 \pm 0.65$ \\
Common & $71.05 \pm 0.73$ & $71.38 \pm 0.75$ & $68.00 \pm 0.62$ & $65.38 \pm 0.86$ & $66.25 \pm 0.61$ \\
CounterFact & $76.25 \pm 0.50$ & $74.82 \pm 0.29$ & $73.34 \pm 0.33$ & $71.71 \pm 0.27$ & $67.68 \pm 0.88$ \\
HateXplain & $74.01 \pm 0.36$ & $73.41 \pm 0.62$ & $70.82 \pm 0.37$ & $66.27 \pm 0.42$ & $64.86 \pm 0.57$ \\
Sarcasm & $89.30 \pm 0.17$ & $89.38 \pm 0.12$ & $86.17 \pm 0.28$ & $87.96 \pm 0.15$ & $81.13 \pm 0.57$ \\
\addlinespace[1pt]
Six-dataset mean & $82.47 \pm 0.16$ & $82.49 \pm 0.10$ & $79.63 \pm 0.11$ & $78.97 \pm 0.16$ & $75.15 \pm 0.30$ \\
\midrule
\multicolumn{6}{l}{\textit{Gemma-7B-it}} \\
STSA & $89.24 \pm 0.27$ & $89.00 \pm 0.12$ & $85.60 \pm 0.12$ & $87.19 \pm 0.32$ & $83.86 \pm 0.16$ \\
Cities & $90.60 \pm 0.50$ & $93.05 \pm 0.29$ & $87.39 \pm 0.49$ & $88.93 \pm 0.44$ & $87.18 \pm 0.83$ \\
Common & $69.49 \pm 0.50$ & $69.71 \pm 0.85$ & $65.28 \pm 0.67$ & $63.23 \pm 0.70$ & $65.81 \pm 0.83$ \\
CounterFact & $71.25 \pm 0.61$ & $69.29 \pm 0.73$ & $68.12 \pm 0.58$ & $65.29 \pm 0.58$ & $66.09 \pm 0.87$ \\
HateXplain & $71.84 \pm 0.55$ & $71.35 \pm 0.74$ & $67.68 \pm 0.38$ & $68.09 \pm 0.49$ & $62.52 \pm 0.73$ \\
Sarcasm & $85.09 \pm 0.30$ & $85.13 \pm 0.25$ & $81.26 \pm 0.48$ & $82.58 \pm 0.24$ & $78.17 \pm 0.93$ \\
\addlinespace[1pt]
Six-dataset mean & $79.58 \pm 0.12$ & $79.59 \pm 0.09$ & $75.89 \pm 0.09$ & $75.88 \pm 0.10$ & $73.94 \pm 0.28$ \\
\midrule
\multicolumn{6}{l}{\textit{Llama-3.1-8B-Instruct}} \\
STSA & $91.92 \pm 0.29$ & $91.65 \pm 0.39$ & $89.73 \pm 0.35$ & $89.97 \pm 0.35$ & $83.76 \pm 0.51$ \\
Cities & $94.41 \pm 0.14$ & $96.79 \pm 0.16$ & $92.69 \pm 0.28$ & $93.72 \pm 0.30$ & $88.44 \pm 0.94$ \\
Common & $71.41 \pm 0.84$ & $71.81 \pm 0.64$ & $67.76 \pm 0.83$ & $65.51 \pm 0.86$ & $66.95 \pm 0.53$ \\
CounterFact & $80.01 \pm 0.35$ & $78.91 \pm 0.46$ & $75.11 \pm 0.54$ & $76.05 \pm 0.51$ & $68.20 \pm 1.25$ \\
HateXplain & $74.96 \pm 0.42$ & $74.25 \pm 0.71$ & $71.66 \pm 0.40$ & $67.60 \pm 0.62$ & $66.63 \pm 0.49$ \\
Sarcasm & $90.09 \pm 0.25$ & $90.08 \pm 0.35$ & $86.98 \pm 0.28$ & $88.83 \pm 0.27$ & $82.74 \pm 0.69$ \\
\addlinespace[1pt]
Six-dataset mean & $83.80 \pm 0.11$ & $83.92 \pm 0.08$ & $80.66 \pm 0.05$ & $80.28 \pm 0.19$ & $76.12 \pm 0.40$ \\
\midrule
\multicolumn{6}{l}{\textit{Llama-3.1-70B-Instruct}} \\
STSA & $93.33 \pm 0.31$ & $92.96 \pm 0.38$ & $91.05 \pm 0.33$ & $92.05 \pm 0.32$ & $82.26 \pm 1.47$ \\
Cities & $96.53 \pm 0.25$ & $97.72 \pm 0.14$ & $95.13 \pm 0.24$ & $96.12 \pm 0.22$ & $93.13 \pm 0.69$ \\
Common & $73.60 \pm 0.92$ & $73.69 \pm 0.83$ & $70.89 \pm 0.80$ & $68.74 \pm 0.85$ & $68.15 \pm 0.50$ \\
CounterFact & $84.65 \pm 0.65$ & $84.01 \pm 0.44$ & $80.57 \pm 0.59$ & $81.90 \pm 0.31$ & $71.17 \pm 1.07$ \\
HateXplain & $77.37 \pm 0.57$ & $76.81 \pm 0.90$ & $74.99 \pm 0.54$ & $70.99 \pm 0.54$ & $68.67 \pm 0.87$ \\
Sarcasm & $93.76 \pm 0.36$ & $93.70 \pm 0.33$ & $91.27 \pm 0.39$ & $93.31 \pm 0.21$ & $88.05 \pm 0.36$ \\
\addlinespace[1pt]
Six-dataset mean & $86.54 \pm 0.07$ & $86.48 \pm 0.13$ & $83.98 \pm 0.02$ & $83.85 \pm 0.08$ & $78.57 \pm 0.47$ \\
\bottomrule
\end{tabular*}
\end{table}
\clearpage

\clearpage
\subsection{Full directional robustness results}
\label{app:robustness-full}

Table~\ref{tab:robustness-full} reports all 18 model--dataset settings and five methods evaluated in Section~\ref{sec:exp_robust}.
For each layer, robustness is the mean absolute pairwise cosine similarity among $K=20$ unit-normalized directions, excluding intercepts.
Max and Mean summarize the complete evaluated layer set; the Median row takes the median of each statistic across the 18 equally weighted settings.
These are descriptive robustness summaries, not uncertainty estimates across five outer split seeds.

\begin{table}[!htbp]
\centering
\caption{\textbf{Complete directional robustness results under $20\%$ training-data ablations.} Max and Mean denote the maximum and mean layerwise robustness, respectively. All values are rounded to two decimal places; a displayed value of $1.00$ does not imply exact unit stability.}
\label{tab:robustness-full}
\small
\setlength{\tabcolsep}{2pt}
\renewcommand{\arraystretch}{1.08}
\begin{tabular*}{\textwidth}{@{\extracolsep{\fill}}l*{10}{c}@{}}
\toprule
 & \multicolumn{2}{c}{\textsc{Raptor}} & \multicolumn{2}{c}{xRFM} & \multicolumn{2}{c}{GCS} & \multicolumn{2}{c}{Fixed-reg} & \multicolumn{2}{c}{CAA} \\
\cmidrule(lr){2-3}\cmidrule(lr){4-5}\cmidrule(lr){6-7}\cmidrule(lr){8-9}\cmidrule(lr){10-11}
Dataset & Max & Mean & Max & Mean & Max & Mean & Max & Mean & Max & Mean \\
\midrule
\multicolumn{11}{l}{\textit{Qwen2.5-3B-Instruct (36 layers)}} \\
STSA & 1.00 & 0.90 & 0.83 & 0.77 & 1.00 & 0.94 & 1.00 & 0.87 & 1.00 & 1.00 \\
Cities & 1.00 & 0.93 & 0.94 & 0.67 & 1.00 & 0.95 & 1.00 & 0.94 & 1.00 & 0.97 \\
Common & 1.00 & 0.94 & 0.72 & 0.69 & 1.00 & 0.96 & 1.00 & 0.88 & 0.99 & 0.99 \\
CounterFact & 1.00 & 0.90 & 0.81 & 0.60 & 1.00 & 0.93 & 1.00 & 0.87 & 1.00 & 0.93 \\
HateXplain & 1.00 & 0.93 & 0.75 & 0.67 & 1.00 & 0.94 & 1.00 & 0.91 & 1.00 & 0.99 \\
Sarcasm & 1.00 & 0.98 & 0.83 & 0.79 & 1.00 & 0.98 & 1.00 & 0.91 & 1.00 & 0.99 \\
\midrule
\multicolumn{11}{l}{\textit{Qwen2.5-32B-Instruct (64 layers)}} \\
STSA & 1.00 & 0.80 & 0.84 & 0.78 & 1.00 & 0.90 & 1.00 & 0.81 & 1.00 & 1.00 \\
Cities & 1.00 & 0.96 & 0.96 & 0.79 & 1.00 & 0.98 & 1.00 & 0.96 & 1.00 & 0.98 \\
Common & 1.00 & 0.92 & 0.72 & 0.71 & 1.00 & 0.97 & 1.00 & 0.80 & 0.99 & 0.99 \\
CounterFact & 1.00 & 0.86 & 0.81 & 0.68 & 1.00 & 0.91 & 1.00 & 0.81 & 1.00 & 0.95 \\
HateXplain & 1.00 & 0.89 & 0.76 & 0.73 & 1.00 & 0.96 & 1.00 & 0.81 & 1.00 & 0.99 \\
Sarcasm & 1.00 & 0.94 & 0.84 & 0.82 & 1.00 & 0.99 & 1.00 & 0.88 & 1.00 & 0.99 \\
\midrule
\multicolumn{11}{l}{\textit{Gemma-7B-it (28 layers)}} \\
STSA & 0.95 & 0.85 & 0.81 & 0.76 & 1.00 & 0.96 & 0.82 & 0.76 & 1.00 & 1.00 \\
Cities & 1.00 & 0.86 & 0.83 & 0.67 & 1.00 & 0.94 & 0.99 & 0.74 & 1.00 & 0.97 \\
Common & 0.99 & 0.82 & 0.71 & 0.69 & 0.99 & 0.89 & 0.88 & 0.70 & 0.99 & 0.99 \\
CounterFact & 0.99 & 0.80 & 0.76 & 0.60 & 1.00 & 0.91 & 0.76 & 0.68 & 1.00 & 0.93 \\
HateXplain & 0.99 & 0.87 & 0.73 & 0.65 & 1.00 & 0.92 & 0.99 & 0.74 & 0.99 & 0.99 \\
Sarcasm & 1.00 & 0.88 & 0.83 & 0.79 & 1.00 & 0.96 & 1.00 & 0.81 & 0.99 & 0.98 \\
\midrule
Median & 1.00 & 0.90 & 0.81 & 0.70 & 1.00 & 0.95 & 1.00 & 0.81 & 1.00 & 0.99 \\
\bottomrule
\end{tabular*}
\end{table}

xRFM retains standardized-space directions, whereas the other methods use raw embedding coordinates.
Exact historical matching of ablation samples across methods remains unverified; comparisons are descriptive.
For Fixed-reg on Gemma-7B-it/STSA, 16 ablation fits at the first evaluated layer reached the iteration limit.
Their finite directions are retained; that layer determines the setting maximum and also contributes to its mean.
These results therefore describe outputs under the fixed optimization budget without asserting convergence of every fit.
\clearpage

\newpage
\clearpage
\section{External assets and licenses}
\label{app:external_assets}

We use only existing public datasets, pretrained models, and previously published baseline methods.
All datasets and methods are credited in the paper, and all pretrained model weights are used under their original access terms.
We do not redistribute pretrained model weights or derived model assets.
Table~\ref{tab:external_assets_licenses} summarizes the external assets used in the experiments and the corresponding license or terms of use available from the original sources.
Where an explicit dataset license could not be verified, this is noted in the table.

We use existing publicly available datasets and pretrained models,
and credit the original datasets and baseline methods in the paper.
Dataset files are converted to CSV for our experimental pipeline
and remain subject to the licenses and terms of their original sources.
Pretrained models are used under their original access terms,
and we do not redistribute their weights.
Table~\ref{tab:external_assets_licenses} summarizes the external datasets
and pretrained models used in our experiments, together with the
license information stated by the cited sources.

\begin{table}[!htb]
\centering
\caption{External datasets and pretrained models used in the experiments,
with license information reported by their sources.}
\label{tab:external_assets_licenses}
\small
\setlength{\tabcolsep}{4pt}
\renewcommand{\arraystretch}{1.08}
\begin{tabular*}{\textwidth}{
    @{\extracolsep{\fill}}
    p{0.23\textwidth}
    p{0.13\textwidth}
    p{0.43\textwidth}
    p{0.14\textwidth}
    @{}
}
\toprule
Asset & Type & License / terms of use & Use in this paper \\
\midrule

STSA / Stanford Sentiment Treebank
& Dataset
& License listed as unknown in the
  \href{https://huggingface.co/datasets/stanfordnlp/sst/blob/main/README.md}
  {SST data card}.
& Probing / steering benchmark \\

\href{https://github.com/saprmarks/geometry-of-truth}{Cities}
& Dataset
& Underlying \href{https://download.geonames.org/export/dump/readme.txt}
  {GeoNames data}: CC BY 4.0.
  No benchmark-specific license stated in the source repository.
& Probing benchmark \\

Common / \href{https://github.com/Algorithmic-Alignment-Lab/CommonClaim}
{CommonClaim}
& Dataset
& Dataset license not specified in the source repository.
& Probing benchmark \\

CounterFact
& Dataset
& Dataset-specific license not stated on the
  \href{https://rome.baulab.info/data/dsets/}
  {official download page}.
& Probing / steering benchmark \\

HateXplain
& Dataset
& CC BY 4.0 in the
  \href{https://huggingface.co/datasets/Hate-speech-CNERG/hatexplain/blob/main/README.md}
  {data-card metadata};
  MIT in the same card's licensing text.
& Probing / steering benchmark \\

Sarcasm News Headlines
& Dataset
& CC BY 4.0, as listed in the
  \href{https://www.kaggle.com/datasets/rmisra/news-headlines-dataset-for-sarcasm-detection}
  {author's Kaggle release}.
& Probing / steering benchmark \\

Llama-3.1-8B-Instruct and Llama-3.1-70B-Instruct
& Pretrained models
& \href{https://huggingface.co/meta-llama/Llama-3.1-8B-Instruct/blob/main/LICENSE}
  {Llama 3.1 Community License Agreement}
  and accompanying Acceptable Use Policy.
& Representation extraction \\

Qwen2.5-3B-Instruct
& Pretrained model
& \href{https://huggingface.co/Qwen/Qwen2.5-3B-Instruct/blob/main/LICENSE}
  {Qwen Research License Agreement}.
& Representation extraction \\

Qwen2.5-7B-Instruct and Qwen2.5-32B-Instruct
& Pretrained models
& \href{https://qwenlm.github.io/blog/qwen2.5/}
  {Apache License 2.0}.
& Representation extraction \\

Gemma-7B-it
& Pretrained model
& \href{https://ai.google.dev/gemma/terms}
  {Gemma Terms of Use}.
& Representation extraction \\

\bottomrule
\end{tabular*}
\end{table}

\newpage
\section*{NeurIPS Paper Checklist}

\begin{enumerate}

\item {\bf Claims}
    \item[] Question: Do the main claims made in the abstract and introduction accurately reflect the paper's contributions and scope?
    \item[] Answer: \answerYes{}
    \item[] Justification: The abstract and introduction state the algorithmic contribution, the evaluation axes, and the high-dimensional theory, and the claims are supported by the experiments in Section~3 and the analysis in Section~4.

\item {\bf Limitations}
    \item[] Question: Does the paper discuss the limitations of the work performed by the authors?
    \item[] Answer: \answerYes{}
    \item[] Justification: Section~5 discusses the main limitations, including the stylized Gaussian theory and the need to extend the analysis to more realistic feature dependencies; the experimental scope and compute behavior are described in Section~3 and Appendix~B.

\item {\bf Theory assumptions and proofs}
    \item[] Question: For each theoretical result, does the paper provide the full set of assumptions and a complete (and correct) proof?
    \item[] Answer: \answerYes{}
    \item[] Justification: The assumptions for the theoretical statements are stated in Section~4, and Appendix~A provides the CGMT-based derivation and references the prior ridge-logistic characterization on which the adapted theorem is based.

    \item {\bf Experimental result reproducibility}
    \item[] Question: Does the paper fully disclose all the information needed to reproduce the main experimental results of the paper to the extent that it affects the main claims and/or conclusions of the paper (regardless of whether the code and data are provided or not)?
    \item[] Answer: \answerYes{}
    \item[] Justification: Sections~2--3 and Appendix~B specify the probe construction, train/validation/test protocol, hyperparameter selection, baselines, steering protocol, and additional diagnostic experiments needed to reproduce the main results.

\item {\bf Open access to data and code}
    \item[] Question: Does the paper provide open access to the data and code, with sufficient instructions to faithfully reproduce the main experimental results, as described in supplemental material?
    \item[] Answer: \answerYes{}
\item[] Justification: Code and instructions for reproducing the main experimental results, including the five-seed benchmark and added baselines, are publicly available at \url{https://github.com/Ziqi-Gao/RAPTOR}. The datasets and pretrained models are available from their original sources.


\item {\bf Experimental setting/details}
    \item[] Question: Does the paper specify all the training and test details (e.g., data splits, hyperparameters, how they were chosen, type of optimizer) necessary to understand the results?
    \item[] Answer: \answerYes{}
    \item[] Justification: The paper specifies the data splits, layerwise training protocol, train-only standardization, ridge grid and validation tuning, baseline protocols, and steering hyperparameters in Sections~2--3 and Appendix~B.

\item {\bf Experiment statistical significance}
    \item[] Question: Does the paper report error bars suitably and correctly defined or other appropriate information about the statistical significance of the experiments?
    \item[] Answer: \answerYes{}
    \item[] Justification: Main probe-accuracy results are summarized over five outer split seeds, with sample standard deviations reported in Appendix~\ref{app:exp}. Appendix~\ref{app:grouped_split} also reports a 95\% confidence interval for a seed-paired \textsc{Raptor}-minus-xRFM comparison. Directional robustness is reported descriptively.

\item {\bf Experiments compute resources}
    \item[] Question: For each experiment, does the paper provide sufficient information on the computer resources (type of compute workers, memory, time of execution) needed to reproduce the experiments?
    \item[] Answer: \answerYes{}
    \item[] Justification: The paper specifies the compute used for the main experiment grid: one NVIDIA H100 GPU was allocated per job and each model--dataset--layer probe fit used one CPU worker. Figure~\ref{fig:cost_dumbbell} reports wall-clock per-layer probe training time across the full grid, and Appendix~\ref{app:steer-protocol} gives the steering-evaluation settings.
    
\item {\bf Code of ethics}
    \item[] Question: Does the research conducted in the paper conform, in every respect, with the NeurIPS Code of Ethics \url{https://neurips.cc/public/EthicsGuidelines}?
    \item[] Answer: \answerYes{}
    \item[] Justification: The work uses existing public models and datasets, does not collect new human-subject data, and is intended for analysis and controlled steering evaluation; we are not aware of any aspect that violates the NeurIPS Code of Ethics.

\item {\bf Broader impacts}
    \item[] Question: Does the paper discuss both potential positive societal impacts and negative societal impacts of the work performed?
    \item[] Answer: \answerYes{}
    \item[] Justification: Section~5 includes a broader-impact discussion: the method is intended to support analysis and controllability of language-model representations, while the same capability could be misused to manipulate model behavior or amplify undesirable concepts. The paper therefore recommends task-specific safety evaluation and monitoring for downstream deployment.
    
\item {\bf Safeguards}
    \item[] Question: Does the paper describe safeguards that have been put in place for responsible release of data or models that have a high risk for misuse (e.g., pre-trained language models, image generators, or scraped datasets)?
    \item[] Answer: \answerNA{}
    \item[] Justification: The paper does not release new pretrained language models, image generators, scraped datasets, or other high-risk assets requiring additional release safeguards.

\item {\bf Licenses for existing assets}
    \item[] Question: Are the creators or original owners of assets (e.g., code, data, models), used in the paper, properly credited and are the license and terms of use explicitly mentioned and properly respected?
    \item[] Answer: \answerNo{}
\item[] Justification:
We cite the original sources of the datasets, pretrained models,
and baseline methods, and summarize the available license and
usage information in Table~\ref{tab:external_assets_licenses}.
Explicit dataset licenses could not be verified for some benchmarks;
missing or inconsistent license information is noted in the table.

\item {\bf New assets}
    \item[] Question: Are new assets introduced in the paper well documented and is the documentation provided alongside the assets?
    \item[] Answer: \answerNA{}
    \item[] Justification: The paper does not introduce or release new datasets, pretrained models, or other standalone assets at submission time; it introduces a method and reports experiments on existing assets.

\item {\bf Crowdsourcing and research with human subjects}
    \item[] Question: For crowdsourcing experiments and research with human subjects, does the paper include the full text of instructions given to participants and screenshots, if applicable, as well as details about compensation (if any)? 
    \item[] Answer: \answerNA{}
    \item[] Justification: The paper does not involve new crowdsourcing experiments or new research with human subjects. It uses existing public benchmark datasets.

\item {\bf Institutional review board (IRB) approvals or equivalent for research with human subjects}
    \item[] Question: Does the paper describe potential risks incurred by study participants, whether such risks were disclosed to the subjects, and whether Institutional Review Board (IRB) approvals (or an equivalent approval/review based on the requirements of your country or institution) were obtained?
    \item[] Answer: \answerNA{}
    \item[] Justification: The paper does not involve new research with human subjects, so IRB or equivalent approval is not applicable.

\item {\bf Declaration of LLM usage}
    \item[] Question: Does the paper describe the usage of LLMs if it is an important, original, or non-standard component of the core methods in this research? Note that if the LLM is used only for writing, editing, or formatting purposes and does \emph{not} impact the core methodology, scientific rigor, or originality of the research, declaration is not required.
    \item[] Answer: \answerYes{}
    \item[] Justification: The paper describes the LLM-as-judge protocol used for the external steering evaluation in Appendix~\ref{app:llm_steering_eval}. LLMs are not part of the proposed probe-training algorithm; any use of LLMs for writing or editing does not affect the core methodology, scientific rigor, or originality of the work.

\end{enumerate}
\end{document}